\documentclass{article}
\usepackage{amssymb}
\usepackage[final]{corl_2025} 

\usepackage{amsmath}
\usepackage{enumitem}
\usepackage{graphicx}
\usepackage{booktabs}
\usepackage{arydshln}
\usepackage{multirow}
\usepackage{tabularx}
\usepackage{makecell}
\usepackage{wrapfig}
\usepackage{xfrac}
\usepackage[ruled,vlined]{algorithm2e}
\usepackage{amsmath}
\usepackage[font=small]{caption}
\usepackage[table]{xcolor} 
\definecolor{headergray}{gray}{0.9}
\definecolor{headergreen}{rgb}{0.89, 0.93, 0.85}
\usepackage{pgfplots}
\pgfplotsset{compat=1.17}
\usepackage{tikz}
\usepackage[most]{tcolorbox}
\usepackage{fvextra}
\usepackage[T1]{fontenc}
\usepackage{geometry}
\DeclareMathOperator*{\argmax}{argmax} 


\newtcolorbox{PromptBlock}[1]{
  colback=gray!5,
  colframe=black,
  title=#1,
  fonttitle=\bfseries,
  arc=2mm,
  enhanced,
  breakable,
  boxrule=0.5pt,
  width=\textwidth,
  left=2mm, right=2mm, top=1mm, bottom=1mm
}

\title{{Search-TTA: A Multimodal Test-Time Adaptation Framework for Visual Search in the Wild} \\ {\large Project Page: \url{https://search-tta.github.io}}}

\author{
    Derek Ming Siang Tan$^{1,4\dagger}$\quad
    Shailesh$^{3}$\quad
    Boyang Liu$^{1}$\quad
    Alok Raj$^{3}$\quad
    Qi Xuan Ang$^{4}$\\
    \textbf{Weiheng Dai}$^{1}$\quad
    \textbf{Tanishq Duhan}$^{1}$\quad
    \textbf{Jimmy Chiun}$^{1}$\quad
    \textbf{Yuhong Cao}$^{1\dagger}$\\
    \textbf{Florian Shkurti}$^{2}$\quad
    \textbf{Guillaume Sartoretti}$^{1}$\\
    $^{1}$National University of Singapore\quad
    $^{2}$University of Toronto\\
    $^{3}$IIT-Dhanbad \quad
    $^{4}$Singapore Technologies Engineering\quad
}

\begin{document}
\renewcommand{\thefootnote}{}
\footnotetext{\footnotesize $^\dagger$Correspondence to: \texttt{derektan@u.nus.edu}, \texttt{caoyuhong@u.nus.edu}}

\maketitle

\vspace{-8mm}
\begin{abstract}

To perform outdoor visual navigation and search, a robot may leverage satellite imagery to generate visual priors.
This can help inform high-level search strategies, even when such images lack sufficient resolution for target recognition.
However, many existing informative path planning or search-based approaches either assume no prior information, or use priors without accounting for how they were obtained.
Recent work instead utilizes large Vision Language Models (VLMs) for generalizable priors, but their outputs can be inaccurate due to hallucination, leading to inefficient search.
To address these challenges, we introduce \textbf{Search-TTA}, a multimodal test-time adaptation framework with a flexible plug-and-play interface compatible with various input modalities (e.g., image, text, sound) and planning methods (e.g., RL-based). 
First, we pretrain a satellite image encoder to align with CLIP's visual encoder to output probability distributions of target presence used for visual search.
Second, our TTA framework dynamically refines CLIP's predictions during search using uncertainty-weighted gradient updates inspired by Spatial Poisson Point Processes.
To train and evaluate Search-TTA, we curate \textbf{AVS-Bench}, a visual search dataset based on internet-scale ecological data containing 380k images and taxonomy data.
We find that Search-TTA improves planner performance by up to 30.0\%, particularly in cases with poor initial CLIP predictions due to domain mismatch and limited training data. 
It also performs comparably with significantly larger VLMs, and achieves zero-shot generalization via emergent alignment to unseen modalities.
Finally, we deploy Search-TTA on a real UAV via hardware-in-the-loop testing, by simulating its operation within a large-scale simulation that provides onboard sensing.

\end{abstract}


\begin{figure*}[h]
    \centering
    \includegraphics[width=\textwidth]{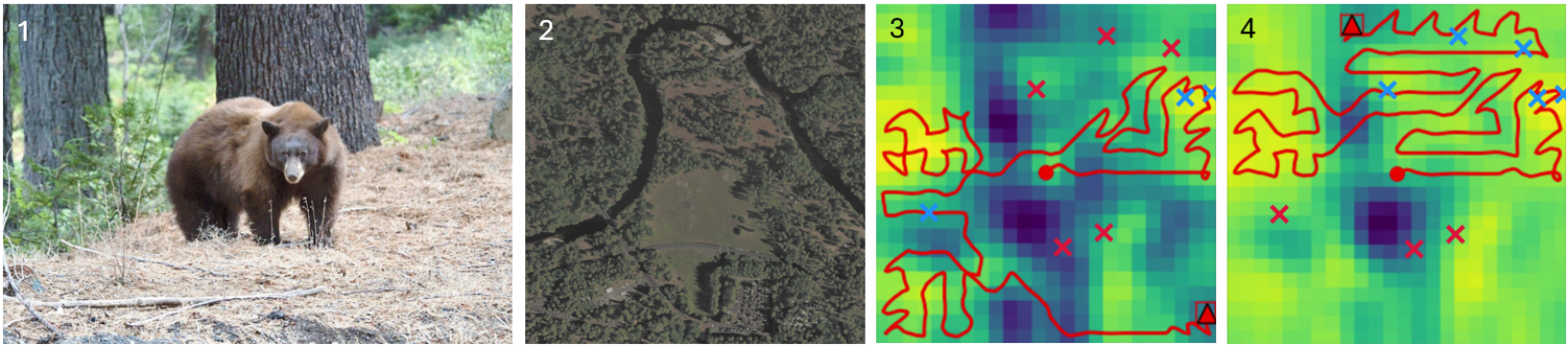}
    \footnotesize
    \caption{\textbf{Visual search for bears by a simulated UAV over Yosemite Valley (2).} (3) Utilizing a poor probability map leads to suboptimal search performance. (4) Test-time adaptation refines the target probability map via regularized gradient updates by incorporating onboard measurements collected during search, guiding the UAV toward denser and less populated vegetation where bears are more likely to be found~\cite{usgs2022naip}~\cite{inat2022inat}.}
    \vspace{-0.3cm}
    \label{fig:visual-search-bear-poster}
\end{figure*}

\section{Introduction}
\vspace{-0.05cm}

Recent advances in visual navigation have leveraged either purely vision-based approaches~\cite{shah2023vint,sridhar2023nomad,bar2024navigation,shah2022viking} or vision–language foundation models (VLN)~\cite{gu2022vision,anderson2018vision,shah2022lmnav} to achieve strong performance and generalization.
VLN approaches are used in Object Goal Navigation (\textit{ObjectNav})~\cite{zeng2020semantic,chaplot2020object,zhai2023peanut}, where robots search for specific household objects indoors.
More broadly, Autonomous Visual Search (AVS) extends to outdoor settings, where robots actively explore natural environments to locate targets of interest, with applications in environmental monitoring~\cite{koreitem2020one}, and search and rescue~\cite{niroui2019deep,wang2024navformer}. 
Outdoor AVS can be particularly challenging due to limited battery life and sensor field-of-view (FOV).
There, one strategy is to extract visual priors from satellite images to direct the search process at a high level, even if the targets cannot be directly seen in these images.
However, many existing informative path planning (IPP) or search-based approaches either avoid using any priors~\cite{vashisth2024deep}, or use them without accounting for how they were obtained~\cite{miller2016ergodic,moon2022tigris}.
Approaches that do generate priors comprise a vision module and a search module. 
The vision module is responsible for processing visual semantics from satellite images and outputs useful priors, either in the form of probability distributions~\cite{sarkar2023partially} or visual embeddings~\cite{sarkar2024visual}.
The planner can then use these inputs to guide the agent towards areas with a higher likelihood of seeing targets, by taking measurements using its higher-resolution sensor.

There are several challenges with generating useful visual priors to guide the search module. 
First, there are limited \textit{in-the-wild} datasets of satellite images with diverse annotated targets that are not directly visible~\cite{thoreau2021sarnet}.
Even if such training data were available, conventional vision models trained on a narrow set of classes~\cite{sarkar2024visual,sarkar2023partially,wang2023spatio} may lack the capacity to reason beyond what is directly observable~\cite{johnson2017clevr}.
Vision-language models (VLMs)~\cite{radford2021learning,li2023blip2,liu2023visual,hurst2024gpt,lai2024lisa}, pretrained on a large corpus of internet-scale data, provide a promising solution to this problem due to their advanced reasoning and generalization abilities.
Unlike conventional vision models, they can reason about correlations between target and semantics of the environment.
Nevertheless, even the best VLMs may generate inaccurate visual outputs (or '\textit{hallucinate}') due to insufficient/inaccurate training data~\cite{Rawte2023hallucination} or when encountering inputs (i.e. satellite images, taxonomies) that are out-of-domain~\cite{yuan2023robust}.  
Over time, these inaccurate predictions persist as VLMs lack a mechanism to correct these errors during search~\cite{yuan2023robust}.

To address these issues, we present \textbf{Search-TTA}, a multimodal test-time adaptation framework that refines a VLM's potentially inaccurate predictions online, using the agent's measurements during AVS.
In this work, we use CLIP~\cite{radford2021learning} as our lightweight VLM, and first align a satellite image encoder to the same representation space~\cite{girdhar2023imagebind,mall2023remote,sastry2025taxabind} a vision encoder through patch-level contrastive learning.
This enables the satellite image encoder to generate a score map by taking the cosine similarity between its per-patch embeddings and the embeddings of other modalities (e.g., ground image, text, sound). 
We then introduce a novel test-time adaptation feedback mechanism to refine CLIP's predictions based on new measurements.
To achieve this, we take inspiration from Spatial Poisson Point Processes~\cite{diggle2013sppp} to perform gradient updates to the satellite image encoder based on past measurements.
We also enhance the loss function with an uncertainty-driven weighting scheme that acts as a regularizer to ensure stable gradient updates. 
To train and evaluate Search-TTA, we curate \textbf{AVS-Bench}, a visual search dataset based on internet-scale ecological data~\cite{van2021inat} comprising satellite images, each with targets and their corresponding ground-level image and taxonomic label (some with sound data). 
It contains \textbf{380k} training and \textbf{8k} validation images (\textit{in-} and \textit{out-domain}).

Search-TTA improves planner performance and score map distribution by up to 30.0\% and 8.5\% respectively, particularly when CLIP predictions are poor due to limited training data and evaluation on \textit{out-domain} taxonomies.
We also demonstrate zero-shot generalization via emergent alignment~\cite{girdhar2023imagebind} with text and sound modalities, without further fine-tuning the satellite image encoder (Fig.~\ref{fig:multimodality_illustration}).

\section{Related Works}
\label{sec:citations}
\vspace{-0.05cm}
\textbf{Visual Navigation:}
There has been significant progress on visual navigation using purely vision~\cite{shah2023vint,sridhar2023nomad,bar2024navigation,shah2022viking} or vision-language foundation models (VLN)~\cite{gu2022vision,anderson2018vision,shah2022lmnav} to achieve high performance and generalization.
VLNs are also used in Object Goal Navigation (\textit{ObjectNav}), where a robot is required to search for objects of interest in indoor household environments.   
Before the emergence of VLNs, these search objects were limited to a closed set~\cite{zeng2020semantic,chaplot2020object,zhai2023peanut}, but more recently, can extend to open sets described via natural language~\cite{gadre2023cow,yu2023leveraging,Yokoyama2024vision,chang2024goat}.
However, outdoor visual search, despite its relevance to tasks like path planning~\cite{tan2024ir}, exploration~\cite{ma2024privileged}, or monitoring~\cite{chiun2025marvel}, remains relatively underexplored, with most prior works only working with closed-set targets~\cite{sarkar2024visual,sarkar2023partially,wang2023spatio}. 
While newer methods leverage foundation models to achieve better results~\cite{sarkar2024gomaa,wang2024navformer}, they tend to have end-to-end architectures and require re-training when the vision backbone or planner changes.
Instead, we focus on a modular approach to connect pre-trained VLMs to off-the-shelf search planners in a flexible manner.

\textbf{Multimodal Learning:}
Ever since the emergence of powerful VLMs~\cite{radford2021learning,li2023blip2,liu2023visual,hurst2024gpt,lai2024lisa}, there has been significant progress in training language foundation models with different modalities, such as audio~\cite{radford2022whisper,Guzhov2022audioclip,pratap2024scaling}, point clouds~\cite{xu2024pointllm,yang2025lidar}, and to output action commands~\cite{kim24openvla,black2024pi0}.
In the remote sensing community, there has been significant interest in training language models with satellite images for semantic segmentation~\cite{ye2025GSNet,trujillano2024image}, visual question-answer~\cite{kuckreja2023geochat,dhakal2024sat2cap} and predictive environmental monitoring~\cite{brown2025alpha,bodnar2025aurora}. 
However, collecting aligned data across multiple modalities remains costly. 
Some works focus on chaining multiple models together to achieve multi-modality~\cite{zeng2023socratic,yang2023mmreact}, but may experience domain mismatch due to their pretraining on different datasets.
Recent efforts align different modalities to a shared representation space~\cite{girdhar2023imagebind,mall2023remote,sastry2025taxabind} to achieve zero-shot generalization via emergent alignment between modalities not jointly present in the training set.
Our work explores this concept by embodying it in the context of visual search that can be prompted by inputs of varying modalities.

\textbf{Online Adaptation:}
Online, or test-time adaptation (TTA), is essential for foundation models facing out-of-domain distributions, and has roots in prior works on continual learning~\cite{wolcyzk2021continual,huang2021continual,meng2025preserving}.
In robotics, online adaptation is associated with meta-learning for few-shot learning~\cite{finn2017model,zhou2020watch} and online adaptation to disturbances in robot dynamics~\cite{finn2019learning}.
Online adaptation is also related to replanning via Chain-of-Thought prompting~\cite{wei2022chain} applied to text~\cite{brohan2023rt2,mu2023embodied} or vision~\cite{huang2022inner,skreta2024replan}, to generate intermediate step-by-step explanations before providing better text/action output.
Other approaches involve direct backpropagation to modify prompts~\cite{lester2021power,liu2021p,shu2022test}, model weights~\cite{zhao2023test,song2023eco}, and to handle dynamic distribution shifts~\cite{yuan2023robust,niu2023towards}.
However, online adaptation with foundation models on satellite images is relatively underexplored.
One example~\cite{elnoor2024robot} uses a robot to navigate the scene using LLM-based traversability estimates, and uses feedback to update its prompts during navigation. 
Conversely, our work explores TTA for visual search using detection measurements to perform weight updates.

\vspace{-0.1cm}
\section{Problem Formulation}
\label{sec:result}
\textbf{Environment:}
We frame AVS as an informative path planning (IPP) problem~\cite{moon2022tigris,vashisth2024deep}, where a robot is tasked with searching for multiple targets over a given satellite image within a time budget. 
First, we consider the search domain over a satellite map $\mathcal{S}$ as a grid map $\mathcal{M}$ composed of $n \times n$ uniform cells, where $\mathcal{M} = \{\psi_1, \psi_2, ...\}$ and $\psi$ represents potential detection viewpoints corresponding to each cell on the map.
We model the target distribution as a subset of grids $\mathcal{M}_t$, where $\mathcal{M}_t \subset \mathcal{M}$. 
Each grid may contain one or more targets, and these target locations are unknown \textit{a-priori}. 

\textbf{Visual Priors:}
We generalize this formulation to accept search queries $G$ of different modalities, such as ground image $G_{i}$, text $G_{t}$, or sound $G_{s}$. 
Pairs of input modalities $(G,\mathcal{S})$ are passed into a vision model to generate visual priors ${p(T \mid G,\mathcal{S})}$ to inform the search process.
Such visual priors can take the form of embeddings in end-to-end frameworks~\cite{sarkar2024visual}, or predicted target probability distributions in frameworks where the vision and search modules are decoupled~\cite{sarkar2023partially}.

\textbf{Target Search:}
Here, the robot is tasked to utilize the visual priors ${p(T \mid G,\mathcal{S})}$ to sequentially explore these cells $\mathcal{M}$ in order to determine the target locations $\mathcal{M}_t$. 
We model our target detection sensor to cover only the grid cell where the robot is currently located $\psi_r$. 
We define the trajectory of viewpoints for the robot $\psi = (\psi_{1}, \psi_{2},\ldots,\psi_{m})$, $\psi_{i} \in \mathcal{M}$. 
The general IPP problem aims to find an optimal trajectory $\psi^*$ in the space of all available trajectories $\Psi$ for maximum gain in some information-theoretic measure, where $\mathrm{I}(\psi)$ is the information gain measured by number of targets found, $\mathrm{C}(\psi)$ is the execution cost measured by steps taken, and $\mathcal{B}$ is the budget constraint.

\vspace{-0.2cm}
\begin{equation} \label{eq2}
    \begin{aligned}
    \psi^* = \argmax_{\psi \in \Psi} \mathrm{I}(\psi) 
    \text{, \hspace{0.25cm} 
    s.t. $\mathrm{C}(\psi) \leq \mathcal{B}$  }
    \end{aligned}
\end{equation}

\section{AVS-Bench Ecological Dataset}
\label{sec:dataset-generation}

\begin{figure*}[t]
    \centering
    \vspace{-0.4cm}
    \includegraphics[width=\textwidth]{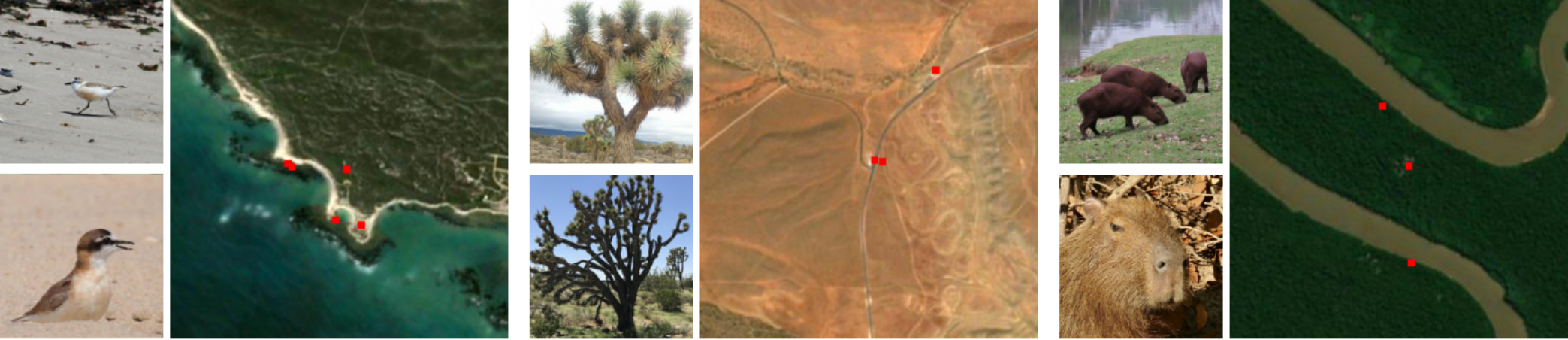}
    \vspace{-0.5cm}
    \footnotesize
    \caption{\textbf{Visual search dataset} The taxonomic targets~\cite{van2021inat} cannot be directly seen on these satellite image~\cite{van2016sentinel}, thus prompting the need to rely on indirect visual cues to achieve efficient search.}
    \vspace{-0.3cm}
    \label{fig:visual-search-dataset}
\end{figure*}

There are limited datasets of satellite images with annotated targets that are not directly visible.
To address this gap, we curate \textbf{AVS-Bench}, a visual search dataset based on internet-scale ecological data.
It comprises Sentinel-2 level 2A satellite images~\cite{van2016sentinel} with unseen taxonomic targets from the iNat-2021 dataset~\cite{van2021inat}, each tagged with ground-level image and taxonomic label (some with sound data).
One advantage of using ecological data is the hierarchical structure of taxonomic labels (seven distinct tiers), which facilitates baseline evaluation across various levels of specificity. 
AVS-Bench is diverse in geography and taxonomies (Appendix~\ref{app:dataset:coverage} \&~\ref{app:dataset:stats}) to reflect \textit{in-the-wild} scenarios.

\textbf{Taxonomic Location Dataset:}
Our goal is to generate a dataset where each image contains multiple target locations for the same taxonomy. 
We begin with the \textit{iSatNat} dataset \cite{sastry2025taxabind} with \textbf{2.7M} satellite images, each matching a ground-level image of a specific taxonomic label.  
We notice a significant amount of overlaps between satellite images, since taxonomies are located at the center of each image.  
Similar to \cite{mall2023remote}, we apply a filter to obtain \textbf{441k} non-overlapping images and store the taxonomy-to-image mappings.  
Thereafter, we store a subset of images with $\geq$ 3 distinct landmarks (to focus on semantic-rich images) and within a range of 3-20 counts of the same taxonomy.
We then split the remaining taxonomies from these filtered images equally into two distinct categories: \textit{in-domain} and \textit{out-domain} taxonomies.
Using these new taxonomy categories, we further split the images into the \textbf{80k} training, \textbf{4k} \textit{in-domain} validation, and \textbf{4k} \textit{out-domain} validation datasets.

\textbf{Taxonomic Score Maps:}
Existing VLMs are trained on large-scale datasets of natural images taken from egocentric viewpoints, and require fine-tuning to perform well on satellite images. However, there are limited remote sensing datasets that correlate segmentation masks with the likelihood of targets.
Although it would be ideal to pretrain our VLMs using the taxonomic location dataset, they only include point locations, and conversion to segmentation masks with likelihood scores is non-trivial. 
We detail our procedures to convert the \textbf{80k} training dataset into score maps in Appendix~\ref{app:dataset:score-map-gen}.

\textbf{Training Dataset Usage:}
We finally train Search-TTA on \textbf{380k} \textit{in-domain} images, obtained from the original \textbf{441k} non-overlapping images after excluding images from the validation sets and keeping only \textit{in-domain} taxonomies.
For our VLM baselines, we train them using our \textbf{80k} score maps.
We use a smaller training dataset for our VLM baselines due to the cost of using GPT4o API to generate score maps, as well as the significant computational resources required to train these large VLMs.
Moreover, these VLMs already have the added advantages of using CLIP as a foundation and being pretrained on much larger datasets.
We include the VLM training details in Appendix~\ref{app:baselines:vlm}.

\section{Search-TTA Framework}
\label{sec:search-tta-framework}

\begin{figure*}[t]  
    \centering
    \vspace{-0.15cm}
    \includegraphics[width=0.93\textwidth]{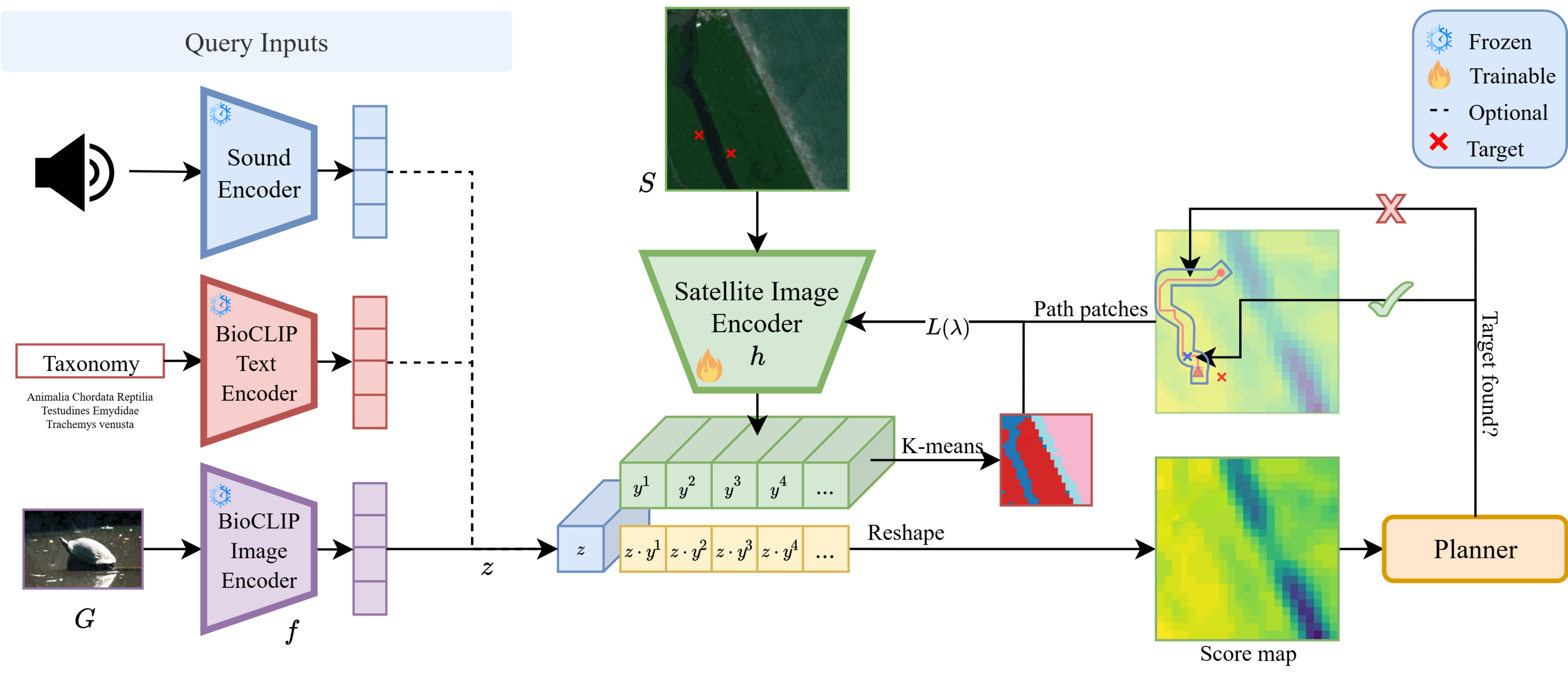}
    \vspace{-0.35cm}
    \footnotesize
    \caption{\textbf{Search-TTA Framework.} During inference, given a satellite image and query inputs of other modalities, a score map is generated to guide the planner towards areas of high probability. The score map (initially poor) will be updated via SPPP gradient updates to the satellite image encoder during the search process.}
    \vspace{-0.5cm}
    \label{fig:tta-framework}
\end{figure*}

We introduce \textbf{Search-TTA}, a multimodal test-time adaptation framework for AVS (Fig.~\ref{fig:tta-framework}), capable of generating and updating probability distributions while collecting measurements during the search process.
We provide visualization and the algorithmic flow in Appendix~\ref{app:search-tta:qualitative} and~\ref{app:search-tta:algorithm} respectively.

\subsection{Multimodal Score Map Generation}
To accept queries of different modalities (e.g. text, image), we need to align their encoder outputs to the same representation space. 
We select the BioCLIP (ViT-B/16)~\cite{stevens2024bioclip} embedding space that is pretrained on the large-scale TreeOfLife dataset~\cite{stevens2024bioclip}, which aligns taxonomy names to ground-level images.
In addition, we train a satellite image encoder by fine-tuning a CLIP (ViT-L/14@336px)~\cite{radford2021learning} image encoder to align with BioCLIP's embedding space. 
We achieve this alignment via the patch-level contrastive loss objective introduced in~\cite{zhang2023patch}, which is a modified version of the standard InfoNCE loss~\cite{oord2018representation} that performs alignment at image level.
Intuitively, this objective aligns the features of the ground images closer with the features of their corresponding patches on the satellite map.  
Through this alignment, we achieve emergent alignment between text and satellite image (Fig.~\ref{fig:multimodality_illustration}).

Consider the pair of modalities ($G$, $\mathcal{S}$), where $G$ denotes the ground-level image and $\mathcal{S}$ denotes the satellite image. 
Given a dataset containing numerous pairs of the two modalities, we organize them into mini-batches $\{g_i, s_i\}_{i=1,...N}$ for training, where $g$ and $s$ refer to the ground images and their corresponding satellite patches respectively. 
We pass the inputs into their respective ground image encoder $f$ and satellite image encoder $h$ , and obtain their normalized embeddings as $z_i = f(g_i) $ and $y_i = h(s_i)$.
Similar to~\cite{mall2023remote}, we remove the pooling layer prior to the final projection layer of the CLIP model to output a per-patch feature vector $z_i[p]$ across the entire satellite image.
We then project $z_i[p]$ from the hidden dimension of 1024 to 512 to match the projection dimension of $y_i$.
Subsequently, we compute the patch-level contrastive loss as such:
\begin{equation}
\mathcal{L}_{\mathcal{G} \rightarrow \mathcal{S}} = 
\frac{1}{N_B}\sum_{i \in N_B} \frac{1}{|G(i)|} 
\sum_{j \in G(i)} -\log \frac{\exp(z_i[p] \cdot y_j / \tau)}{\sum_{n \in N_B} \exp(z_i[p] \cdot y_n / \tau)}
\vspace{-0.05cm}
\end{equation}
where $N_B$ is the total number of pairs in the mini-batch, $G(i)$ the set of ground images with the same species category as that of the $i^{th}$ satellite image, and $\tau$ the temperature parameter.
Similarly, we define the patch-level contrastive loss $\mathcal{L}_{\mathcal{S} \rightarrow \mathcal{G}}$, and take the average of $\mathcal{L}_{\mathcal{G} \rightarrow \mathcal{S}}$ and $\mathcal{L}_{\mathcal{S} \rightarrow \mathcal{G}}$ as the final loss.
Using AVS-Bench detailed in Sec.~\ref{sec:dataset-generation}, we fine-tune the CLIP model with two NVIDIA A6000 GPUs, which took 3.5 days before convergence.
During training, we update the weights of the satellite image encoder while keeping BioCLIP frozen. 
During inference, we generate the 24 $\times$ 24 probability distribution by taking the cosine distance between the query input features with all satellite image patch features (Fig.~\ref{fig:tta-framework}).
Further training details can be found in Appendix~\ref{app:search-tta:training-sat-encoder}.

\subsection{Search Planners}
\vspace{-0.025cm}
Search-TTA is designed to be adaptable to different types of search planners. 
This ranges from conventional methods such as Information Surfing (IS)~\cite{lanillos2014multi} to Deep Reinforcement Learning (RL)~\cite{cao2023ariadne} methods. 
In principle, the Search-TTA framework can be applied to other types of planners as long as they can reasonably utilize the probability map from the vision model to inform their search strategies.
We detail the performance of Search-TTA with various search planners in Section~\ref{sec:expt-tta-planners}.

\subsection{Test-Time Adaptation Feedback Loop}
\vspace{-0.025cm}
One of the key features of Search-TTA is its ability to refine probability distribution outputs from the vision model based on collected measurements.
This process can be broken down into two stages.

\textbf{K-means Clustering of CLIP Embeddings:}
Before the start of each search episode, we perform k-means clustering of the per-patch satellite image features $z_i[p]$ to generate clusters of embeddings that are semantically similar~\cite{ma2024mode}. 
These clusters correspond to regions used in the modified Spatial Poisson Point Process (SPPP)~\cite{diggle2013sppp} loss function below. This avoids the 
need to rely on external segmentation tools~\cite{kirillov2023segment, ren2024grounded}.
In this work, we choose the best $k$ by taking the average of the silhouette score criterion~\cite{shahapure2020silhouette} and the elbow criterion~\cite{marutho2018elbow}. 
More details can be found in Appendix~\ref{app:search-tta:kmeans}.

\textbf{SPPP-based Online Adaptation:}
During search, the robot uses sensors to detect targets and collects feedback to refine CLIP's probability predictions.
We adapt the Negative Log-Likelihood loss function from inhomogeneous SPPP to perform gradient updates for CLIP. SPPP is a statistical model to describe the frequency of scattered points in space with state-dependent intensity functions $\lambda$. 
While SPPP uses absolute $\lambda$ values, CLIP's likelihood outputs can be approximated as normalized $\lambda$ values across all regions. 
We can thus adapt SPPP's update function and apply it to CLIP.

Note that the vanilla loss function~\cite{diggle2013sppp} does not work as a test-time update mechanism, because it was designed to regress SPPPs over a large batch of available data during training. 
In our case, the robot begins with no prior knowledge of the targets' locations and has to perform detection along the search process.  
For scenarios with sparse targets, the modes of the CLIP probability distribution may quickly collapse because the robot will collect many negative measurements before finding a first target. 
To address this, we introduce an uncertainty-driven weighting scheme that acts as a regularizer to the loss function, where $p$ and $n$ are the positive and negative measurements collected.
\vspace{-0.1cm}
\vspace{-2.5mm}
\begin{equation}
L(\lambda) = \sum_{i=1}^{p} \left( \alpha_{\text{pos},i} \right) \log \lambda(x_i) 
- \sum_{j=1}^{n} \left( \alpha_{\text{neg},j} \right) \lambda(x_j) \, dx.
\vspace{-1.5mm}
\end{equation}
Intuitively, we do not want to significantly reduce the probability of a semantic region after only a few negative measurements, since they may not accurately represent the overall distribution. 
Hence, we scale negative measurements with the coefficient $\alpha_{\text{neg},j}$ based on how much of the corresponding semantic region has been covered.
Similar to the concept of focal loss~\cite{lin2018focal}, we introduce an exponent $\gamma$ (=2 in practice) to give less weight to measurements in regions that are largely uncovered. 
In practice, we use $\alpha_{\text{neg,j}} =  \min \left(\beta \left ( O_r / L_r \right)^{\gamma},\ 1 \right)$, where $O_r$ is the number of patches observed in region $r$ and $L_r$ is the number of patches in that region.
For $\alpha_{\text{pos},i}$, we find that keeping it constant (=4 in our case) works well in practice. 
Note that we reset the satellite encoder weights back to the base weights before running TTA updates. 
We employ a learning rate schedule that gradually increases our gradient updates to maintain effective learning when more measurements are collected~\cite{goyal2018accurate}.

\vspace{-0.1cm}
\section{Experiments}
\label{sec:experiments}
\vspace{-0.1cm}

\begin{table}[t] 
\vspace{-4mm} 
\scriptsize
\centering  
\caption{Evaluating TTA on different planners (CLIP Vision Model), on Out-domain taxonomies}
\resizebox{\columnwidth}{!}{
\setlength{\tabcolsep}{3pt}
\begin{tabular}{lccccccc@{}p{0.4em}@{}ccccccc}

\toprule
\multirow{3}{*}{\raisebox{-0.3\normalbaselineskip}[0pt][0pt]{\textbf{Planner Type}}} 
    & \multicolumn{7}{>{\columncolor{headergreen}}c}{\textbf{$\mathcal{B}=256$}} 
    & \multicolumn{1}{c}{} 
    & \multicolumn{7}{>{\columncolor{headergreen}}c}{\textbf{$\mathcal{B}=384$}} \\
\addlinespace[1mm]
    & \multicolumn{3}{c}{\textbf{Found (\%)} $\uparrow$} 
    & \multicolumn{3}{c}{\textbf{RMSE (\%)} $\downarrow$}
    & \multirow{2}{*}{\raisebox{-2.0ex}{\makecell[c]{\textbf{Steps  $\downarrow$}\\\tiny\textbf{(First tgt)}}}}
    & 
    & \multicolumn{3}{c}{\textbf{Found (\%)} $\uparrow$} 
    & \multicolumn{3}{c}{\textbf{RMSE (\%)} $\downarrow$}
    & \multirow{2}{*}{\raisebox{-2.0ex}{\makecell[c]{\textbf{Steps  $\downarrow$}\\\tiny\textbf{(First tgt)}}}} \\
\cmidrule(lr){2-4} \cmidrule(lr){5-7} \cmidrule(lr){10-12} \cmidrule(lr){13-15}
\noalign{\vskip -0.4ex}
    & {\tiny\textbf{\textit{All}}} & {\tiny\textbf{\textit{Bot. 5\%}}} & {\tiny\textbf{\textit{Bot. 2\%}}}
    & {\tiny\textbf{\textit{First}}} & {\tiny\textbf{\textit{Mid}}} & {\tiny\textbf{\textit{Last}}}
    & 
    &
    & {\tiny\textbf{\textit{All}}} & {\tiny\textbf{\textit{Bot. 5\%}}} & {\tiny\textbf{\textit{Bot. 2\%}}}
    & {\tiny\textbf{\textit{First}}} & {\tiny\textbf{\textit{Mid}}} & {\tiny\textbf{\textit{Last}}}
    & \\
\noalign{\vskip -0.3ex} 

\midrule
RL (TTA)~\cite{cao2023ariadne} 
    & 60.8 & 31.7 & \textbf{30.7} 
    & 54.2 & 53.2 & \textbf{51.2} 
    & 86.5
    & 
    & 79.6 & 58.9 & \textbf{56.1} 
    & 54.2 & 52.7 & \textbf{47.0} 
    & 103.8 \\

RL (no TTA)~\cite{cao2023ariadne} 
    & 58.5 & 23.1 & \textbf{16.0} 
    & 54.2 & 54.2 & \textbf{54.2} 
    & 88.2
    & 
    & 77.1 & 44.8 & \textbf{36.1} 
    & 54.2 & 54.2 & \textbf{54.2} 
    & 107.5 \\

\midrule

IS (TTA)~\cite{lanillos2014multi} 
    & 53.9 & 24.2 & \textbf{22.2} 
    & 54.2 & 52.7 & \textbf{51.3} 
    & 92.7
    & 
    & 74.0 & 46.1 & \textbf{40.6} 
    & 54.2 & 52.7 & \textbf{47.2} 
    & 114.8 \\

IS (no TTA)~\cite{lanillos2014multi} 
    & 51.2 & 19.1 & \textbf{12.9} 
    & 54.2 & 54.2 & \textbf{54.2} 
    & 92.3
    & 
    & 71.9 & 32.3 & \textbf{23.8} 
    & 54.2 & 54.2 & \textbf{54.2} 
    & 115.6 \\

\midrule

Lawnmower~\cite{choset2001coverage} 
    & 41.7 & -- & -- 
    & -- & -- & -- 
    & 118.8
    &
    & 74.2 & -- & -- 
    & -- & -- & -- 
    & 157.6 \\
    
\bottomrule
\end{tabular}
}
\label{tab:expt-tta-planners}
\vspace{-6mm}
\end{table}

The main objective of our experiments is to test Search-TTA's ability to enhance AVS performance, while providing a flexible plug-and-play interface compatible with various query modalities and planning methods. 
We run our experiments using the AVS-Bench validation datasets.
Unless mentioned otherwise, we discretize the search space to 24$\times$24 grids, randomize start positions, use ground-level images as our query modality, and perform TTA updates every 20 steps or whenever targets are found.
We aim to answer the following:
\vspace{-0.2cm}
\begin{enumerate}
    \item How effective is Search-TTA in improving AVS performance of different planners?
    \vspace{-0.5mm}
    \item How does Search-TTA compare to other VLMs and AVS frameworks? 
    \vspace{-0.5mm}
    \item Does Search-TTA exhibit zero-shot generalization to different unseen modality inputs?
\end{enumerate}

\subsection{Effectiveness of TTA on Different Planners}
\label{sec:expt-tta-planners}
We compare Search-TTA with an Attention-based RL planner~\cite{cao2023ariadne} (pretrained on score maps from Sec.~\ref{sec:dataset-generation}) and a greedy IS planner~\cite{lanillos2014multi}, while using Lawnmower~\cite{choset2001coverage} as a no-priors baseline.
Since these planners input binary visited states as observations, running them without TTA is similar to IPP approaches with binary sensors~\cite{moon2022tigris,vashisth2024deep}.
We provide further details on these baselines in Appendix~\ref{app:baselines:planner}
In Table~\ref{tab:expt-tta-planners}, we report the average percentage of targets found, Root Mean Squared Error (RMSE) between CLIP predictions and ground truth score maps (from Sec.~\ref{sec:dataset-generation}), and the steps taken to reach the first target, all within 256 steps and averaged across all \textbf{4k} \textit{out-domain} validation images. 
To further determine Search-TTA's effectiveness, we recorded the targets found given poor CLIP predictions, namely in
the bottom 5\% and 2\% percentiles in terms of CLIP prediction quality.
To do so, we take the average scores of the pixels where the targets are located on the predicted score map and deem a CLIP prediction to be poor if most targets are located in low-scoring regions.
Fig.~\ref{fig:tta-diff-trendline-256} and Fig.~\ref{fig:tta-diff-trendline-384} reflect that TTA performance gain is most significant in the bottom percentiles across all planners, indicating its ability to correct poor initial score maps for \textit{in-the-wild} data.

Our results show a general improvement across all metrics with Search-TTA.
We note the most significant improvement of 20.0\% in the bottom 2\% scoring CLIP predictions using the RL planner when $\mathcal{B}=384$. 
Furthermore, we observe a decreasing trend in RMSE of up to 7.2\%, which indicates that predicted score maps become increasingly accurate with TTA iterations.
We note similar trends when evaluated on \textit{in-domain} taxonomies (Table~\ref{tab:expt-tta-planners-384steps}), with RMSE improvements up to 8.5\%.

\begin{figure}[t]
    \centering
    \vspace{-0.4cm}
    \begin{minipage}[t]{0.58\textwidth}
        \centering
        \scriptsize
        \captionof{table}{Comparing vision models (Found \%) $\uparrow$}
        \vspace{-2mm}
        \setlength{\tabcolsep}{0pt}
        \begin{tabularx}{\linewidth}{
            >{\raggedright\arraybackslash\hspace{0pt}}p{0.40\linewidth}
            @{\hspace{2mm}}                                       
            >{\centering\arraybackslash}X
            >{\centering\arraybackslash}X
            >{\centering\arraybackslash}X
            >{\centering\arraybackslash}X}
        \toprule
        \multirow{2}{*}{\textbf{Vision Models}}
        & \multicolumn{2}{c}{\textbf{In-domain}}
        & \multicolumn{2}{c}{\textbf{Out-domain}} \\
        \cmidrule(lr){2-3} \cmidrule(lr){4-5}
        & $\mathcal{B}=256$ & $\mathcal{B}=384$
        & $\mathcal{B}=256$ & $\mathcal{B}=384$ \\
        \midrule
        CLIP (TTA)~\cite{stevens2024bioclip} & \textbf{57.4} & 76.1 & \textbf{60.8} & \textbf{79.6} \\
        CLIP (no TTA)~\cite{stevens2024bioclip} & 56.6 & 75.5 & 58.5 & 77.1 \\
        LISA~\cite{lai2024lisa} & 57.1 & \textbf{76.9} & 58.4 & 77.8 \\
        LLM-Seg~\cite{wang2024llm} & 52.6 & 71.6 & 54.4 & 73.3 \\
        Qwen2+GroundedSAM~\cite{wang2024qwen2,ren2024grounded} & 51.9 & 72.0 & 55.2 & 74.2 \\
        LLaVA+GroundedSAM~\cite{liu2023visual,ren2024grounded} & 51.7 & 71.6 & 54.6 & 73.5 \\
        \bottomrule
        \end{tabularx}
        \label{tab:expt-vision-models}
    \end{minipage}
    \hspace{2mm}
    \begin{minipage}[t]{0.38\textwidth}
        \centering
        \scriptsize
        \captionof{figure}{VLM inference time}
        \vspace{-2mm}
        \scalebox{0.8}{
            \begin{tikzpicture}
\begin{axis}[
    ybar stacked,
    bar width=0.8cm,
    width=7cm,
    height=5cm,
    ylabel={Inference Time (s)},
    ymin=0,
    ymax=2.5,
    ytick={0,0.5,1.0,1.5,2.0},
    xtick={1,2,3,4,5},
    xticklabels={,,,,},  
    xmin=0.3,
    xmax=5.7,
    xlabel style={font=\normalsize},
    ylabel style={font=\normalsize},
    yticklabel style={font=\normalsize},
    xticklabel style={font=\normalsize},
    ymajorgrids=true,
    grid style={gray!30},
    clip=false,
    axis x line*=bottom,
    axis y line*=left,
    legend style={
        at={(0.02,0.98)},
        anchor=north west,
        legend columns=1,
        row sep=0.02cm,
        draw=none,
        fill=none,
        font=\scriptsize,
        cells={anchor=west},
    },
    area legend,
]

\addplot[
    fill=blue!40,
    draw=none,
    nodes near coords={0.11},
    nodes near coords style={font=\scriptsize, anchor=south, yshift=-0.16cm},
    point meta=explicit symbolic,
] coordinates {
    (1,0.11)
    (2,0)
    (3,0)
    (4,0)
    (5,0)
};
\addlegendentry{CLIP (Base)}

\addplot[
    fill=blue!20,
    draw=none,
    nodes near coords={0.15},
    nodes near coords style={font=\scriptsize, anchor=south, yshift=-0.16cm},
    point meta=explicit symbolic,
] coordinates {
    (1,0.15) [0.26]
    (2,0) []
    (3,0) []
    (4,0) []
    (5,0) []
};
\addlegendentry{CLIP (TTA)}

\addplot[
    fill=green!45!brown,
    draw=none,
    nodes near coords={0.42},
    nodes near coords style={font=\scriptsize, anchor=south, yshift=-0.21cm},
    point meta=explicit symbolic,
] coordinates {
    (1,0) []
    (2,0.42) [0.42]
    (3,0) []
    (4,0) []
    (5,0) []
};
\addlegendentry{LISA}

\addplot[
    fill=purple!50,
    draw=none,
    nodes near coords={0.40},
    nodes near coords style={font=\scriptsize, anchor=south, yshift=-0.20cm},
    point meta=explicit symbolic,
] coordinates {
    (1,0) []
    (2,0) []
    (3,0.40) [0.40]
    (4,0) []
    (5,0) []
};
\addlegendentry{LLM-Seg}

\addplot[
    fill=orange!60,
    draw=none,
    nodes near coords={1.91},
    nodes near coords style={font=\scriptsize, anchor=center},
    point meta=explicit symbolic,
] coordinates {
    (1,0) []
    (2,0) []
    (3,0) []
    (4,1.91) [1.91]
    (5,0) []
};
\addlegendentry{Qwen2+GSAM}

\addplot[
    fill=red!50,
    draw=none,
    nodes near coords={3.48},
    nodes near coords style={font=\scriptsize, anchor=center},
    point meta=explicit symbolic,
] coordinates {
    (1,0) []
    (2,0) []
    (3,0) []
    (4,0) []
    (5,2.48) [2.48]
};
\addlegendentry{LLaVA+GSAM}

\end{axis}
\end{tikzpicture}
        }
        \label{fig:inference_speed}
    \end{minipage}
    \vspace{-7mm}
\end{figure}

\subsection{Comparison with Baselines}
\label{sec:expt-baselines}

\textbf{Varying Vision Model:}
We evaluate the effectiveness of Search-TTA's CLIP vision backbone by replacing it with different state-of-the-art VLMs.  
We modify LISA and LLM-Seg to improve their performance for AVS, and fine-tune them with the score maps from Sec.~\ref{sec:dataset-generation}.
More details about these VLMs, training setup, and hyper-parameters can be found in Appendix~\ref{app:baselines:vlm}.

We run all of these baseline VLMs with our RL planner, and record the targets found (Table~\ref{tab:expt-vision-models}) and inference time (Fig.~\ref{fig:inference_speed}).
Our results indicate that CLIP with TTA generally outperforms all baselines across different budgets except for \textit{in-domain} data when $\mathcal{B}=384$.
In addition, we attribute LLM-Seg's poor performance to limitations in its training and output (training on binary masks only, and discretizing scores in its output maps).
We also note that the fully decoupled baselines perform poorly, likely because Qwen-7B, LLaVA-13B, and GroundedSAM are not fine-tuned with remote sensing data.
Note that CLIP has the fastest inference time on an A5000 GPU (TTA every 20 steps).
We additionally measure the time taken for CLIP inference and TTA on an edge device (Orin AGX) to be 0.14s and 0.37s respectively (Table~\ref{tab:search-tta-time-analysis}), highlighting its suitability for on-board deployment.

\setlength{\intextsep}{3.0pt} 
\begin{wraptable}{r}{0.5\textwidth} 
    \centering
    \scriptsize
    \caption{Comparing AVS frameworks (Found \%)} 
    \vspace{-2mm}
    \setlength{\tabcolsep}{0pt} 
    \begin{tabularx}{\linewidth}{
        >{\raggedright\arraybackslash\hspace{0pt}}p{0.3\linewidth} 
        >{\centering\arraybackslash}X    
        >{\centering\arraybackslash}X  
        >{\centering\arraybackslash}X   
        >{\centering\arraybackslash}X}
    \toprule
    \multirow{2}{*}{\textbf{Frameworks}} 
    & \multicolumn{2}{c}{\textbf{Charadriiformes (In)}} 
    & \multicolumn{2}{c}{\textbf{Columbiformes (Out)}} \\
    \cmidrule(lr){2-3} \cmidrule(lr){4-5}
    & $\mathcal{B}=256$ & $\mathcal{B}=384$ 
    & $\mathcal{B}=256$ & $\mathcal{B}=384$ \\
    \midrule
    CLIP+RL (TTA) & \textbf{60.3} & \textbf{79.7} & \textbf{62.9} & \textbf{82.2} \\
    CLIP+RL (no TTA) & 58.6 & 77.5 & 61.0 & 78.4 \\
    PSVAS~\cite{sarkar2023partially} & 53.0 & 68.5 & 60.3 & 75.0 \\
    VAS~\cite{sarkar2024visual} & 49.5 & 66.2 & 55.7 & 73.3 \\
    Lawnmower~\cite{choset2001coverage} & 41.4 & 72.0 & 38.1 & 74.5 \\
    \bottomrule
    \end{tabularx}
    \label{tab:expt-avs-frameworks}
\end{wraptable}

\textbf{AVS Baselines:}
We evaluate the effectiveness of Search-TTA by comparing it against existing AVS baselines (VAS and PSVAS) in the remote sensing domain. 
While VAS utilizes end-to-end reinforcement learning, PSVAS decouples vision and search models while introducing test-time adaptation. 
Likewise, we fine-tune these baselines with the score maps from Sec.~\ref{sec:dataset-generation}, and modify them to enhance their performance.
More details can be found in Appendix~\ref{app:baselines:avs-frameworks}.

We evaluate our approach and the baselines using images with two different sub-classes of birds (\textit{Animalia Chordata Aves}) as search targets, namely \textit{Charadriiformes} and \textit{Columbiformes}, which are more likely to be found along shorelines and on urban areas respectively.
As seen in Table~\ref{tab:expt-avs-frameworks}, CLIP with TTA and without TTA both significantly outperform the AVS baselines by up to 13.5\% and 11.3\% respectively.  
We note similar trends when evaluated on \textit{Animalia Chordata Reptilia Squamata} (Appendix~\ref{app:add-expts:baselines}).
Although Lawnmower outperforms VAS and PSVAS when $\mathcal{B}=$384, VAS and PSVAS can find the first target more quickly by performing a more targeted search (Table~\ref{tab:expt-search-frameworks-v2}).

\subsection{Multimodal Inputs}
\label{sec:expt-multimodal-inputs}
\vspace{-0.5mm}

We evaluate the generalization ability of Search-TTA to previously unseen input modalities (Fig.~\ref{fig:tta-framework}). 
To achieve this, we input the full taxonomic name into the CLIP text encoder, obtaining query text embeddings that are used similarly to the ground image embeddings.
We run these experiments over our \textit{in-domain} validation images (Table~\ref{tab:expt-multimodal}), and note the performance gap of at most 0.9\%.
Separately, we fine-tune and evaluate a sound encoder~\cite{wu2024clap} using the \textit{quad-modal} split of AVS-Bench, achieving a performance gap of at most 2.4\%.
This indicates successful zero-shot generalization via emergent alignment to text and sound modality although we did not fine-tune the satellite image encoder with text/sound data.
We provide more details on our sound dataset in Appendix~\ref{app:dataset:sound-dataset-gen}.

\begin{table}[t]
    \centering
    \vspace{-0.65cm}
    \begin{minipage}[t]{0.57\textwidth}
        \centering
        \scriptsize
        \captionof{table}{Zero-shot generalization (Found \%)}
        \renewcommand{\arraystretch}{0.9}
        \begin{tabularx}{\linewidth}{
            >{\raggedright\arraybackslash}p{0.16\linewidth}
            @{} 
            >{\centering\arraybackslash}p{0.16\linewidth}
            @{\hspace{1.5mm}}                                     
            >{\centering\arraybackslash}X 
            >{\centering\arraybackslash}X 
            >{\centering\arraybackslash}X
            >{\centering\arraybackslash}X 
        }
        \toprule
        \multirow{3}{*}{\shortstack[l]{\textbf{Input}\\\textbf{Modality}}}
        & \multirow{3}{*}{\shortstack{\textbf{Dataset}\\\textbf{Size}}}
        & \multicolumn{2}{c}{\textbf{$\mathcal{B} = 256$}}
        & \multicolumn{2}{c}{\textbf{$\mathcal{B} = 384$}} \\
        \cmidrule(lr){3-4} \cmidrule(lr){5-6}
        & & \textbf{TTA} & \hspace*{-0.35em}\textbf{No~TTA}
          & \textbf{TTA} & \hspace*{-0.35em}\textbf{No~TTA} \\
        \midrule
        Image & 4k & \textbf{57.4} & 56.6 & 76.1 & 75.5 \\
        Text  & 4k & \textbf{56.7} & 55.9 & 75.2 & 74.7 \\
        \midrule
        Image & 460 & 56.2 & 55.0 & 75.1 & \textbf{74.5} \\
        Text  & 460 & 56.9 & 55.7 & \textbf{76.5} & \textbf{74.7} \\
        Sound & 460 & 54.5 & 54.0 & \textbf{75.1} & 73.2 \\
        \bottomrule
        \end{tabularx}
        \label{tab:expt-multimodal}
    \end{minipage}
    \hspace{2mm}
    \begin{minipage}[t]{0.38\textwidth}
        \centering
        \scriptsize
        \captionof{figure}{Multimodal Alignment}
        \includegraphics[width=1.0\textwidth]{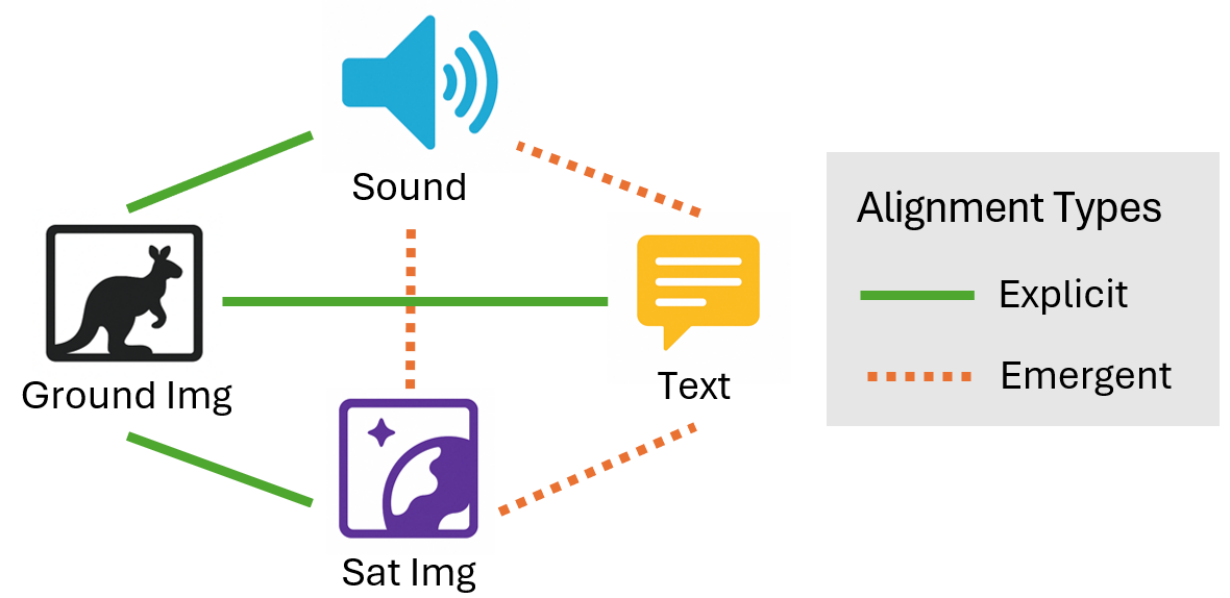}
        \label{fig:multimodality_illustration}
    \end{minipage}
    \vspace{-9.5mm}
\end{table}

\subsection{Ablation Studies}
\vspace{-0.5mm}

\textbf{Scaling of Training Dataset:}
We analyze the impact of training dataset size for our satellite image encoder on TTA performance improvement.
In Fig.~\ref{fig:dataset_scaling_figure}, we observe that models trained on smaller datasets tend to benefit more from TTA.
In Table~\ref{tab:expt-vary-clip-train-ds}, we report up to a 30.0\% increase in targets found with the 80k dataset (bot. 2\%).
This highlights Search-TTA's ability to improve poorer score maps generated by models with limited training data through stronger online corrections.

\textbf{SPPP Loss Coefficient:}
We investigate the effects of the vanilla log-likelihood loss function and hyperparameter tuning in Appendix~\ref{app:search-tta:sppp}. 
When we remove the negative weighting coefficient ($\gamma=0$) or the relative weighting factor ($\beta=1$), we observe poorer performance of up to 8.1\%.

\textbf{Varying TTA methodology:}
We explore the effectiveness of our TTA methodology compared to prompt learning~\cite{shu2022test} and text-based TTA~\cite{huang2022inner} in Appendix~\ref{app:add-expts:tta-methods}.
We notice that our SPPP-based formulation outperforms prompt learning in terms of targets found (Table~\ref{tab:expt-prompt-learning}), and outperforms text-based TTA in terms of consistency in score map improvements during search (Table~\ref{tab:expt-text-tta}).

\subsection{Evaluation on Hardware}
\label{sec:expt-real-robot}
\vspace{-0.5mm}

\begin{figure}[t]
    \centering
    \vspace{-0.45cm}
    \begin{minipage}[t]{0.35\textwidth}
        \centering
        \scriptsize
        \captionof{figure}{Dataset Scaling (Bot. 5\%)}
        \vspace{-0.2cm}
        \scalebox{0.5}{
            \begin{tikzpicture}
\begin{axis}[
    width=11cm,
    height=6cm,
    axis lines=left,
    xlabel={Dataset Size},
    ylabel={TTA Performance Gain (\%)},
    xlabel style={font=\large, yshift=-2pt},
    ylabel style={font=\large},
    xtick={80000, 200000, 380000},
    xticklabels={80k, 200k, 380k},
    tick label style={font=\large},
    xmin=60000, xmax=400000,
    ymin=0, ymax=32,
    scaled x ticks=false,
    legend style={draw=none, fill=white, font=\normalsize, at={(0.98,0.98)}, anchor=north east, legend columns=1},
    legend cell align={left},
    tick align=outside,
    tick style={black},
    grid=major,
]

\addplot[
    color=green!60!black,
    mark=diamond*,
    very thick,
    mark size=2.5pt,
] coordinates {
    (80000,11.3)
    (200000,4.3)
    (380000,1.4)
};
\addlegendentry{Val In-Domain (256 Steps)}

\addplot[
    color=blue,
    mark=square*,
    very thick,
    mark size=2.5pt,
] coordinates {
    (80000,17.9)
    (200000,7.1)
    (380000,3.1)
};
\addlegendentry{Val In-Domain (384 Steps)}

\addplot[
    color=orange,
    mark=triangle*,
    very thick,
    mark size=2.5pt,
] coordinates {
    (80000, 16.4)
    (200000, 7.4)
    (380000, 8.9)
};
\addlegendentry{Val Out-Domain (256 Steps)}

\addplot[
    color=red,
    mark=*,
    very thick,
    mark size=2.5pt,
] coordinates {
    (80000,25.3)
    (200000,14.4)
    (380000,14.2)
};
\addlegendentry{Val Out-Domain (384 Steps)}

\end{axis}
\end{tikzpicture}
        }
        \label{fig:dataset_scaling_figure}
    \end{minipage}
    \hspace{2.5mm}
    \begin{minipage}[t]{0.62\textwidth}
        \centering
        \scriptsize
        \captionof{figure}{AVS with Crazyflie drone (Perception in Gazebo)}
        \vspace{-0.2cm}
        \includegraphics[width=0.95\textwidth]{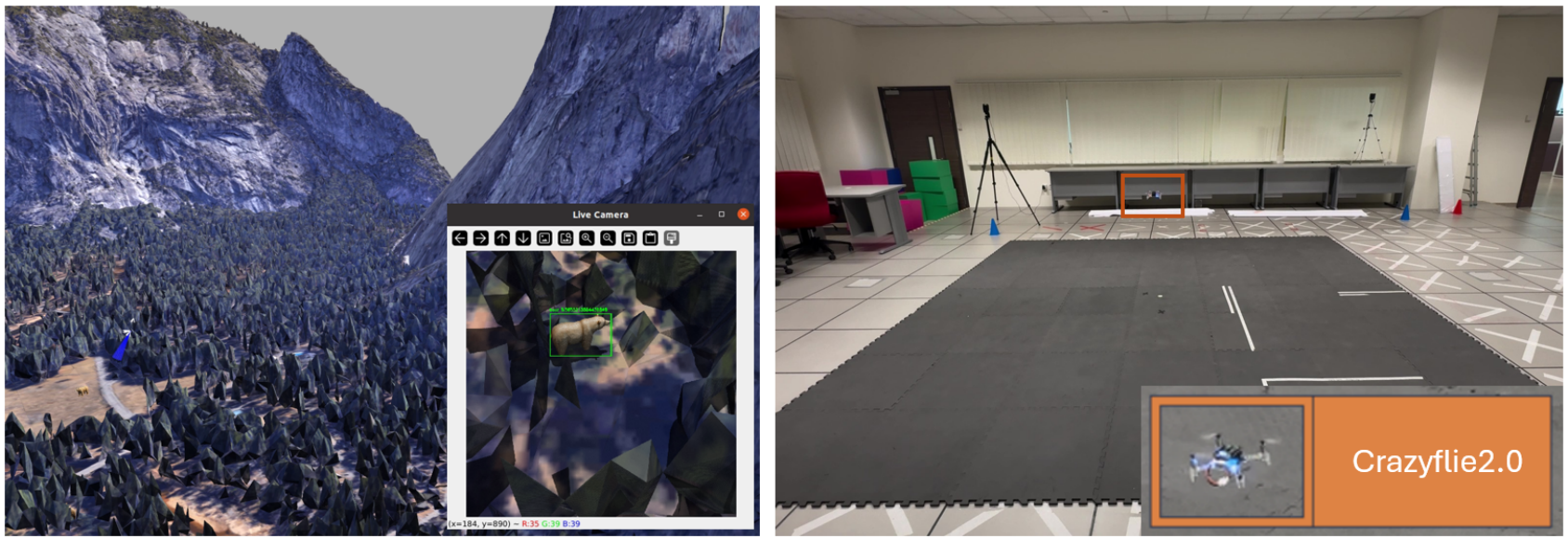}
        \label{fig:crazyfile}
    \end{minipage}
    \vspace{-7mm}
\end{figure}

We carried out hardware-in-the-loop experiments to validate Search-TTA's performance in locating black bears in Yosemite Valley (Fig.~\ref{fig:crazyfile}). 
We deploy a Crazyflie 2.1 drone operating within a 4m $\times$ 4m mockup arena (17 $\times$ 17 grids) with external localization.
Concurrently, we launch a ROS2 drone simulator~\cite{simulator2022tum} within a Yosemite Valley 3D model~\cite{google2022gearth} in Gazebo, from which we obtain onboard measurements from the drone's downward-facing camera (YOLO11x~\cite{glenn2024yolo11} for bear detection).

We conducted one experiment each for scenarios with and without TTA ($\mathcal{B}=300s$), using a NAIP~\cite{usgs2022naip} satellite image of the operating area and an image of a bear sighting from iNaturalist~\cite{inat2022inat} as inputs.
From Fig.~\ref{fig:visual-search-bear-poster}, we note that 5 targets were found with TTA, compared to 3 targets found without TTA.
This is likely due to the inaccurate initial score map output by CLIP, where portions of the densely forested areas were underestimated with low probabilities.
With TTA, the detection of the first bear significantly corrected this initial distribution, guiding the robot to explore the dense forested areas and thus find more targets.
More experimental details can be found in Appendix~\ref{app:add-expts:baselines}.

\section{Conclusion}
\label{sec:conclusion}
\vspace{-0.1cm}
This paper addresses the challenges of autonomous outdoor visual search where targets cannot be seen directly from satellite images.
We introduce \textbf{Search-TTA}, a test-time adaptation framework that enhances potentially inaccurate CLIP predictions.
Our contributions include curating \textbf{AVS-Bench}, an internet-scale dataset with unseen taxonomic targets, enabling multimodal query through alignment to a common representation space, and proposing a novel TTA mechanism inspired by Spatial Poisson Point Processes.
Search-TTA significantly improves planner performance by up to 30.0\%, and performs comparably with significantly larger and state-of-the-art VLMs.
We also demonstrate zero-shot generalization to text and sound modality without additional fine-tuning.
We hope that our research will inspire future work in visual search and ecological conservation.

\section{Limitations and Future Work}
\vspace{-0.1cm}

\textbf{Sensor Model:}
There are a limited number of existing works dealing with AVS over satellite images. 
In order to fairly compare with ~\cite{sarkar2024visual, sarkar2023partially}, we decided to remain consistent with their problem statement, and naturally inherit some of the limitations in their formulation:
\begin{itemize}[leftmargin=*]
\setlength{\itemsep}{0pt}
    \vspace{-0.2cm}
    \item We assume that our sensor model has very narrow field of view. As a result, search performance can be highly stochastic, making performance improvements quite marginal (especially when averaged over a large validation set), as robots can easily miss targets by a small margin.
    \item We assume that our sensor model is perfect and binary. This does not account for detection uncertainty or false positive/negative measurements, which would require robots to re-visit specific areas to confirm their beliefs~\cite{vashisth2024deep,moon2022tigris}. Future work will look at extending our TTA method to handle more complex and realistic search formulations. 
    \item We assume that our sensor model is capable of detecting targets which may be hidden in areas occluded from direct satellite imagery (e.g. dense forest, water surfaces etc.). Future work will consider other sensor modalities (e.g. thermal camera, camera traps etc.) or other types of robots (e.g. unmanned ground/underwater vehicles) to model realistic detection constraints. 
\end{itemize}

\textbf{Beyond Visual Semantics:}
Search-TTA is effective in drawing connections between the target taxonomy and visual semantics, but does not currently consider other essential factors when determining the likelihood of their whereabouts, such as relationships between the different landmarks, interactions between different species, sources of food/danger, etc. Future work will extend the reasoning of our VLM to allow for such deeper, multi-faceted reasoning.

\textbf{Multi-Target Search:}
Adapting Search-TTA to simultaneously search for multiple target types presents significant challenges related to catastrophic forgetting~\cite{hu2018overcoming} during gradient updates when different target types are encountered sequentially. 
Future work will focus on developing continual learning methodologies (e.g. batch normalization or importance sampling~\cite{yuan2023robust}) that enable our framework to maintain performance across previously learned targets while adapting to new ones.

\acknowledgments{ This work was supported by Singapore Technologies Engineering Ltd, under the Economic Development Board - Industrial Postgraduate Program (Project No. 2022-2130).
We would like to extend our gratitude to Kimberly Fornace (School of Public Health) and Adrian Loo Hock Beng (Faculty of Science, Center for Nature-based Climate Solutions) from the National University of Singapore for the helpful conversations on ecological conservation that shaped the early stages of this paper.
}

\clearpage

\bibliography{ref}  

@misc{thoreau2021sarnet,
    title={SaRNet: A Dataset for Deep Learning Assisted Search and Rescue with Satellite Imagery}, 
    author={Michael Thoreau and Frazer Wilson},
    year={2021},
    eprint={2107.12469},
    archivePrefix={arXiv},
    primaryClass={eess.IV}
}

@inproceedings{shah2023vint,
  title     = {Vi{NT}: A Foundation Model for Visual Navigation},
  author    = {Dhruv Shah and Ajay Sridhar and Nitish Dashora and Kyle Stachowicz and Kevin Black and Noriaki Hirose and Sergey Levine},
  booktitle = {7th Annual Conference on Robot Learning},
  year      = {2023},
}

@article{sridhar2023nomad,
    author  = {Ajay Sridhar and Dhruv Shah and Catherine Glossop and Sergey Levine},
    title   = {{NoMaD: Goal Masked Diffusion Policies for Navigation and Exploration}},
    journal = {arXiv pre-print},
    year    = {2023},
  }

@misc{bar2024navigation,
    title={Navigation World Models}, 
    author={Amir Bar and Gaoyue Zhou and Danny Tran and Trevor Darrell and Yann LeCun},
    year={2024},
    eprint={2412.03572},
    archivePrefix={arXiv},
    primaryClass={cs.CV},
}

@inproceedings{shah2022viking, 
    author    = {Dhruv Shah and Sergey Levine}, 
    title     = {{ViKiNG: Vision-Based Kilometer-Scale Navigation with Geographic Hints}}, 
    booktitle = {Proceedings of Robotics: Science and Systems}, 
    year      = {2022},
}

@inproceedings{gu2022vision,
    title = "Vision-and-Language Navigation: A Survey of Tasks, Methods, and Future Directions",
    author = "Gu, Jing and Stefani, Eliana and Wu, Qi and Thomason, Jesse and Wang, Xin",
    booktitle = "Proceedings of the 60th Annual Meeting of the Association for Computational Linguistics (Volume 1: Long Papers)",
    year = "2022",
    pages = "7606--7623",
    doi = "10.18653/v1/2022.acl-long.524",
}

@INPROCEEDINGS{anderson2018vision,
  author={Anderson, Peter and Wu, Qi and Teney, Damien and Bruce, Jake and Johnson, Mark and Sünderhauf, Niko and Reid, Ian and Gould, Stephen and van den Hengel, Anton},
  booktitle={2018 IEEE/CVF Conference on Computer Vision and Pattern Recognition}, 
  title={Vision-and-Language Navigation: Interpreting Visually-Grounded Navigation Instructions in Real Environments}, 
  year={2018},
  volume={},
  number={},
  pages={3674-3683},
  doi={10.1109/CVPR.2018.00387}}

@inproceedings{shah2022lmnav,
    title={{LM}-Nav: Robotic Navigation with Large Pre-Trained Models of Language, Vision, and Action},
    author={Dhruv Shah and Blazej Osinski and Brian Ichter and Sergey Levine},
    booktitle={6th Annual Conference on Robot Learning},
    year={2022},
}

@INPROCEEDINGS{zeng2020semantic,
  author={Zeng, Zhen and Röfer, Adrian and Jenkins, Odest Chadwicke},
  booktitle={2020 IEEE International Conference on Robotics and Automation (ICRA)}, 
  title={Semantic Linking Maps for Active Visual Object Search}, 
  year={2020},
  volume={},
  number={},
  pages={1984-1990},
  doi={10.1109/ICRA40945.2020.9196830}
}

@inproceedings{chaplot2020object,
  title={Object Goal Navigation using Goal-Oriented Semantic Exploration},
  author={Chaplot, Devendra Singh and Gandhi, Dhiraj and
            Gupta, Abhinav and Salakhutdinov, Ruslan},
  booktitle={In Neural Information Processing Systems},
  year={2020}
}

@inproceedings{zhai2023peanut,
  title={{PEANUT}: Predicting and Navigating to Unseen Targets},
  author={Zhai, Albert J and Wang, Shenlong},
  booktitle={ICCV},
  year={2023}
}

@INPROCEEDINGS{gadre2023cow,
  author={Gadre, Samir Yitzhak and Wortsman, Mitchell and Ilharco, Gabriel and Schmidt, Ludwig and Song, Shuran},
  booktitle={2023 IEEE/CVF Conference on Computer Vision and Pattern Recognition (CVPR)}, 
  title={CoWs on Pasture: Baselines and Benchmarks for Language-Driven Zero-Shot Object Navigation}, 
  year={2023},
  volume={},
  number={},
  pages={23171-23181},
  doi={10.1109/CVPR52729.2023.02219}
}

@INPROCEEDINGS{yu2023leveraging,
  author={Yu, Bangguo and Kasaei, Hamidreza and Cao, Ming},
  booktitle={2023 IEEE/RSJ International Conference on Intelligent Robots and Systems (IROS)}, 
  title={L3MVN: Leveraging Large Language Models for Visual Target Navigation}, 
  year={2023},
  volume={},
  number={},
  pages={3554-3560},
  doi={10.1109/IROS55552.2023.10342512}
}

@INPROCEEDINGS{Yokoyama2024vision,
  author={Yokoyama, Naoki and Ha, Sehoon and Batra, Dhruv and Wang, Jiuguang and Bucher, Bernadette},
  booktitle={2024 IEEE International Conference on Robotics and Automation (ICRA)}, 
  title={VLFM: Vision-Language Frontier Maps for Zero-Shot Semantic Navigation}, 
  year={2024},
  volume={},
  number={},
  pages={42-48},
  doi={10.1109/ICRA57147.2024.10610712}
}

@inproceedings{chang2024goat,
    author = {Chang, Matthew and Gervet, Theophile and Khanna, Mukul and Yenamandra, Sriram and Shah, Dhruv and Min, So Yeon and Shah, Kavit and Paxton, Chris and Gupta, Saurabh and Batra, Dhruv and Mottaghi, Roozbeh and Malik, Jitendra and Chaplot, Devendra},
    year = {2024},
    pages = {},
    title = {GOAT: GO to Any Thing},
    doi = {10.15607/RSS.2024.XX.073}
}

@INPROCEEDINGS{wang2023spatio,
  author={Wang, Yizhuo and Wang, Yutong and Cao, Yuhong and Sartoretti, Guillaume},
  booktitle={2023 IEEE/RSJ International Conference on Intelligent Robots and Systems (IROS)}, 
  title={Spatio-Temporal Attention Network for Persistent Monitoring of Multiple Mobile Targets}, 
  year={2023},
  volume={},
  number={},
  pages={3903-3910},
  doi={10.1109/IROS55552.2023.10341674}
}

@ARTICLE{wang2024navformer,
  author={Wang, Haitong and Tan, Aaron Hao and Nejat, Goldie},
  journal={IEEE Robotics and Automation Letters}, 
  title={NavFormer: A Transformer Architecture for Robot Target-Driven Navigation in Unknown and Dynamic Environments}, 
  year={2024},
  volume={9},
  number={8},
  pages={6808-6815},
  doi={10.1109/LRA.2024.3412638}
}

@inproceedings{li2023blip2,
  title={{BLIP-2:} Bootstrapping Language-Image Pre-training with Frozen Image Encoders and Large Language Models}, 
  author={Junnan Li and Dongxu Li and Silvio Savarese and Steven Hoi},
  year={2023},
  booktitle={ICML},
}

@misc{radford2022whisper,
  doi = {10.48550/ARXIV.2212.04356},
  author = {Radford, Alec and Kim, Jong Wook and Xu, Tao and Brockman, Greg and McLeavey, Christine and Sutskever, Ilya},
  title = {Robust Speech Recognition via Large-Scale Weak Supervision},
  publisher = {arXiv},
  year = {2022}
}

@INPROCEEDINGS{Guzhov2022audioclip,
  author={Guzhov, Andrey and Raue, Federico and Hees, Jörn and Dengel, Andreas},
  booktitle={ICASSP 2022 - 2022 IEEE International Conference on Acoustics, Speech and Signal Processing (ICASSP)}, 
  title={Audioclip: Extending Clip to Image, Text and Audio}, 
  year={2022},
  volume={},
  number={},
  pages={976-980},
  doi={10.1109/ICASSP43922.2022.9747631}
}

@article{pratap2024scaling,
    author = {Pratap, Vineel and Tjandra, Andros and Shi, Bowen and Tomasello, Paden and Babu, Arun and Kundu, Sayani and Elkahky, Ali and Ni, Zhaoheng and Vyas, Apoorv and Fazel-Zarandi, Maryam and Baevski, Alexei and Adi, Yossi and Zhang, Xiaohui and Hsu, Wei-Ning and Conneau, Alexis and Auli, Michael},
    title = {Scaling speech technology to 1,000+ languages},
    year = {2024},
    publisher = {JMLR.org},
    volume = {25},
    number = {1},
    journal = {J. Mach. Learn. Res.},
}

@inproceedings{xu2024pointllm,
  title={PointLLM: Empowering Large Language Models to Understand Point Clouds},
  author={Xu, Runsen and Wang, Xiaolong and Wang, Tai and Chen, Yilun and Pang, Jiangmiao and Lin, Dahua},
  booktitle={ECCV},
  year={2024}
}

@article{yang2025lidar,
    title={LiDAR-LLM: Exploring the Potential of Large Language Models for 3D LiDAR Understanding}, 
    volume={39}, 
    DOI={10.1609/aaai.v39i9.33001}, 
    journal={Proceedings of the AAAI Conference on Artificial Intelligence}, 
    author={Yang, Senqiao and Liu, Jiaming and Zhang, Renrui and Pan, Mingjie and Guo, Ziyu and Li, Xiaoqi and Chen, Zehui and Gao, Peng and Li, Hongsheng and Guo, Yandong and Zhang, Shanghang}, 
    year={2025}, 
    pages={9247-9255} 
}

@article{kim24openvla,
    title={OpenVLA: An Open-Source Vision-Language-Action Model},
    author={{Moo Jin} Kim and Karl Pertsch and Siddharth Karamcheti and Ted Xiao and Ashwin Balakrishna and Suraj Nair and Rafael Rafailov and Ethan Foster and Grace Lam and Pannag Sanketi and Quan Vuong and Thomas Kollar and Benjamin Burchfiel and Russ Tedrake and Dorsa Sadigh and Sergey Levine and Percy Liang and Chelsea Finn},
    journal = {arXiv preprint arXiv:2406.09246},
    year={2024},
}

@misc{black2024pi0,
      title={$\pi_0$: A Vision-Language-Action Flow Model for General Robot Control}, 
      author={Kevin Black and Noah Brown and Danny Driess and Adnan Esmail and Michael Equi and Chelsea Finn and Niccolo Fusai and Lachy Groom and Karol Hausman and Brian Ichter and Szymon Jakubczak and Tim Jones and Liyiming Ke and Sergey Levine and Adrian Li-Bell and Mohith Mothukuri and Suraj Nair and Karl Pertsch and Lucy Xiaoyang Shi and James Tanner and Quan Vuong and Anna Walling and Haohuan Wang and Ury Zhilinsky},
      year={2024},
}

@article{trujillano2024image,
    author = {Trujillano, Fedra and Jiménez, Gabriel and Manrique Valverde, Edgar Joao and Kahamba, Najat and Okumu, Fredros and Apollinaire, Nombre and Carrasco-Escobar, Gabriel and Barrett, Brian and Fornace, Kimberly},
    year = {2024},
    month = {05},
    pages = {},
    title = {Using image segmentation models to analyse high-resolution earth observation data: new tools to monitor disease risks in changing environments},
    volume = {23},
    journal = {International Journal of Health Geographics},
    doi = {10.1186/s12942-024-00371-w}
}

@article{kuckreja2023geochat,
  title={GeoChat: Grounded Large Vision-Language Model for Remote Sensing},
  author={Kuckreja, Kartik and Danish, Muhammad S. and Naseer, Muzammal and Das, Abhijit and Khan, Salman and Khan, Fahad S.},
  journal={The IEEE/CVF Conference on Computer Vision and Pattern Recognition},
  year={2024}
}

@inproceedings{dhakal2024sat2cap,
  title={Sat2cap: Mapping fine-grained textual descriptions from satellite images},
  author={Dhakal, Aayush and Ahmad, Adeel and Khanal, Subash and Sastry, Srikumar and Kerner, Hannah and Jacobs, Nathan},
  booktitle={IEEE/ISPRS Workshop: Large Scale Computer Vision for Remote Sensing (EARTHVISION)},
  pages={533--542},
  year={2024}
}

@article{brown2025alpha,
    author = {Brown, Christopher and Kazmierski, Michal and Pasquarella, Valerie and Rucklidge, William and Samsikova, Masha and Zhang, Chenhui and Shelhamer, Evan and Lahera, Estefania and Wiles, Olivia and Ilyushchenko, Simon and Gorelick, Noel and Zhang, Lihui and Alj, Sophia and Schechter, Emily and Askay, Sean and Guinan, Oliver and Moore, Rebecca and Boukouvalas, Alexis and Kohli, Pushmeet},
    year = {2025},
    title = {AlphaEarth Foundations: An embedding field model for accurate and efficient global mapping from sparse label data},
    doi = {10.48550/arXiv.2507.22291}
}

@article{bodnar2025aurora,
    title = {A Foundation Model for the Earth System},
    author = {Cristian Bodnar and Wessel P. Bruinsma and Ana Lucic and Megan Stanley and Anna Allen and Johannes Brandstetter and Patrick Garvan and Maik Riechert and Jonathan A. Weyn and Haiyu Dong and Jayesh K. Gupta and Kit Thambiratnam and Alexander T. Archibald and Chun-Chieh Wu and Elizabeth Heider and Max Welling and Richard E. Turner and Paris Perdikaris},
    journal = {Nature},
    year = {2025},
    doi = {10.1038/s41586-025-09005-y},
}

@inproceedings{finn2017model,
author = {Finn, Chelsea and Abbeel, Pieter and Levine, Sergey},
title = {Model-agnostic meta-learning for fast adaptation of deep networks},
year = {2017},
booktitle = {Proceedings of the 34th International Conference on Machine Learning - Volume 70},
pages = {1126–1135}
}

@inproceedings{zeng2023socratic,
    title={Socratic Models: Composing Zero-Shot Multimodal Reasoning with Language},
    author={Andy Zeng and Maria Attarian and brian ichter and Krzysztof Marcin Choromanski and Adrian Wong and Stefan Welker and Federico Tombari and Aveek Purohit and Michael S Ryoo and Vikas Sindhwani and Johnny Lee and Vincent Vanhoucke and Pete Florence},
    booktitle={The Eleventh International Conference on Learning Representations},
    year={2023},
}

@article{yang2023mmreact,
  author      = {Zhengyuan Yang and Linjie Li and Jianfeng Wang and Kevin Lin and Ehsan Azarnasab and Faisal Ahmed and Zicheng Liu and Ce Liu and Michael Zeng and Lijuan Wang},
  title       = {MM-REACT: Prompting ChatGPT for Multimodal Reasoning and Action},
  publisher   = {arXiv},
  year        = {2023},
}

@inproceedings{wolcyzk2021continual,
 author = {Wo\l czyk, Maciej and Zaj\k{a}c, Micha\l  and Pascanu, Razvan and Kuci\'{n}ski, \L ukasz and Mi\l o\'{s}, Piotr},
 booktitle = {Advances in Neural Information Processing Systems},
 pages = {28496--28510},
 title = {Continual World: A Robotic Benchmark For Continual Reinforcement Learning},
 volume = {34},
 year = {2021}
}

@article{meng2025preserving,
  title={Preserving and combining knowledge in robotic lifelong reinforcement learning},
  author={Meng, Yuan and Bing, Zhenshan and Yao, Xiangtong and Chen, Kejia and Huang, Kai and Gao, Yang and Sun, Fuchun and Knoll, Alois},
  journal={Nature Machine Intelligence},
  pages={1--14},
  year={2025},
  publisher={Nature Publishing Group UK London}
}

@INPROCEEDINGS{huang2021continual,
  author={Huang, Yizhou and Xie, Kevin and Bharadhwaj, Homanga and Shkurti, Florian},
  booktitle={2021 IEEE International Conference on Robotics and Automation (ICRA)}, 
  title={Continual Model-Based Reinforcement Learning with Hypernetworks}, 
  year={2021},
  volume={},
  number={},
  pages={799-805},
  doi={10.1109/ICRA48506.2021.9560793}
}

@inproceedings{wei2022chain,
author = {Wei, Jason and Wang, Xuezhi and Schuurmans, Dale and Bosma, Maarten and Ichter, Brian and Xia, Fei and Chi, Ed H. and Le, Quoc V. and Zhou, Denny},
title = {Chain-of-thought prompting elicits reasoning in large language models},
year = {2022},
isbn = {9781713871088},
booktitle = {Proceedings of the 36th International Conference on Neural Information Processing Systems}
}

@inproceedings{brohan2023rt2,
    title={RT-2: Vision-Language-Action Models Transfer Web Knowledge to Robotic Control},
    author={Anthony Brohan and Noah Brown and Justice Carbajal and Yevgen Chebotar and Xi Chen and Krzysztof Choromanski and Tianli Ding and Danny Driess and Avinava Dubey and Chelsea Finn and Pete Florence and Chuyuan Fu and Montse Gonzalez Arenas and Keerthana Gopalakrishnan and Kehang Han and Karol Hausman and Alex Herzog and Jasmine Hsu and Brian Ichter and Alex Irpan and Nikhil Joshi and Ryan Julian and Dmitry Kalashnikov and Yuheng Kuang and Isabel Leal  and Lisa Lee and Tsang-Wei Edward Lee and Sergey Levine and Yao Lu and Henryk Michalewski and Igor Mordatch and Karl Pertsch and Kanishka Rao and Krista Reymann and Michael Ryoo and Grecia Salazar and Pannag Sanketi and Pierre Sermanet and Jaspiar Singh and Anikait Singh and Radu Soricut and Huong Tran and Vincent Vanhoucke and Quan Vuong and Ayzaan Wahid and Stefan Welker and Paul Wohlhart and  Jialin Wu and Fei Xia and Ted Xiao and Peng Xu and Sichun Xu and Tianhe Yu and Brianna Zitkovich},
    booktitle={arXiv preprint arXiv:2307.15818},
    year={2023}
}

@inproceedings{mu2023embodied,
    author = {Mu, Yao and Zhang, Qinglong and Hu, Mengkang and Wang, Wenhai and Ding, Mingyu and Jin, Jun and Wang, Bin and Dai, Jifeng and Qiao, Yu and Luo, Ping},
    title = {EmbodiedGPT: vision-language pre-training via embodied chain of thought},
    year = {2023},
    booktitle = {Proceedings of the 37th International Conference on Neural Information Processing Systems},
}

@inproceedings{huang2022inner,
    title={Inner Monologue: Embodied Reasoning through Planning with Language Models},
    author={Wenlong Huang and Fei Xia and Ted Xiao and Harris Chan and Jacky Liang and Pete Florence and Andy Zeng and Jonathan Tompson and Igor Mordatch and Yevgen Chebotar and Pierre Sermanet and Noah Brown and Tomas Jackson and Linda Luu and Sergey Levine and Karol Hausman and Brian Ichter},
    booktitle={arXiv preprint arXiv:2207.05608},
    year={2022}
}

@misc{skreta2024replan,
      title={RePLan: Robotic Replanning with Perception and Language Models}, 
      author={Marta Skreta and Zihan Zhou and Jia Lin Yuan and Kourosh Darvish and Alán Aspuru-Guzik and Animesh Garg},
      year={2024},
}

@INPROCEEDINGS{song2023eco,
  author={Song, Junha and Lee, Jungsoo and Kweon, In So and Choi, Sungha},
  booktitle={2023 IEEE/CVF Conference on Computer Vision and Pattern Recognition (CVPR)}, 
  title={EcoTTA: Memory-Efficient Continual Test-Time Adaptation via Self-Distilled Regularization}, 
  year={2023},
  pages={11920-11929},
  doi={10.1109/CVPR52729.2023.01147}
}

@inproceedings{niu2023towards,
  title={Towards Stable Test-Time Adaptation in Dynamic Wild World},
  author={Niu, Shuaicheng and Wu, Jiaxiang and Zhang, Yifan and Wen, Zhiquan and Chen, Yaofo and Zhao, Peilin and Tan, Mingkui},
  booktitle = {Internetional Conference on Learning Representations},
  year = {2023}
}

@misc{elnoor2024robot,
  title={Robot Navigation Using Physically Grounded Vision-Language Models in Outdoor Environments}, 
  author={Mohamed Elnoor and Kasun Weerakoon and Gershom Seneviratne and Ruiqi Xian and Tianrui Guan and Mohamed Khalid M Jaffar and Vignesh Rajagopal and Dinesh Manocha},
  year={2024},
}

@misc{google2022gearth,
  author       = {Google},
  title        = {Google Earth Engine},
}

@misc{inat2022inat,
  author       = {iNaturalist},
  title        = {iNaturalist},
  url          = {https://www.inaturalist.org}, 
}

@misc{van2021inat,
  author={Van Horn, Grant and Mac Aodha, Oisin},
  title={iNat Challenge 2021 - FGVC8},
  publisher={Kaggle},
  year={2021},
  url={https://kaggle.com/competitions/inaturalist-2021}
}

@misc{usgs2022naip,
  author       = {U.S.G.S.},
  title        = {National agriculture imagery program (naip)},
  year         = {2022},
}

@inproceedings{ye2025GSNet,
  title={Towards Open-Vocabulary Remote Sensing Image Semantic Segmentation},
  author={Ye, Chengyang and Zhuge, Yunzhi and Zhang, Pingping},
  booktitle={Proceedings of the AAAI Conference on Artificial Intelligence},
  year={2025}
}

@inproceedings{finn2019learning,
    title={Learning to Adapt in Dynamic, Real-World Environments through Meta-Reinforcement Learning},
    author={Ignasi Clavera and Anusha Nagabandi and Simin Liu and Ronald S. Fearing and Pieter Abbeel and Sergey Levine and Chelsea Finn},
    booktitle={International Conference on Learning Representations},
    year={2019},
}

@inproceedings{zhou2020watch,
    title={Watch, Try, Learn: Meta-Learning from Demonstrations and Rewards},
    author={Allan Zhou and Eric Jang and Daniel Kappler and Alex Herzog and Mohi Khansari and Paul Wohlhart and Yunfei Bai and Mrinal Kalakrishnan and Sergey Levine and Chelsea Finn},
    booktitle={International Conference on Learning Representations},
    year={2020},
}

@INPROCEEDINGS{cao2023ariadne,
  author={Cao, Yuhong and Hou, Tianxiang and Wang, Yizhuo and others},
  booktitle={2023 IEEE ICRA}, 
  title={ARiADNE: A Reinforcement learning approach using Attention-based Deep Networks for Exploration}, 
  year={2023},
}

@book{diggle2013sppp,
    author = {Diggle, Peter},
    year = {2013},
    month = {07},
    pages = {1-264},
    title = {Statistical analysis of spatial and spatio-temporal point patterns, third edition},
    isbn = {9780429098093},
    journal = {Statistical Analysis of Spatial and Spatio-Temporal Point Patterns, Third Edition},
    doi = {10.1201/b15326}
}

@misc{lin2018focal,
      title={Focal Loss for Dense Object Detection}, 
      author={Tsung-Yi Lin and Priya Goyal and Ross Girshick and Kaiming He and Piotr Dollár},
      year={2018},
      eprint={1708.02002},
      archivePrefix={arXiv},
      primaryClass={cs.CV},
}

@INPROCEEDINGS{shahapure2020silhouette,
  title={Cluster Quality Analysis Using Silhouette Score}, 
  author={Shahapure, Ketan Rajshekhar and Nicholas, Charles},
  booktitle={2020 IEEE 7th International Conference on Data Science and Advanced Analytics (DSAA)}, 
  pages={747-748},
  year={2020},
}

@INPROCEEDINGS{marutho2018elbow,
  title={The Determination of Cluster Number at k-Mean Using Elbow Method and Purity Evaluation on Headline News}, 
  author={Marutho, Dhendra and Hendra Handaka, Sunarna and Wijaya, Ekaprana and Muljono},
  booktitle={2018 International Seminar on Application for Technology of Information and Communication}, 
  pages={533-538},
  year={2018},
}

@article{lanillos2014multi,
    title = {Multi-UAV target search using decentralized gradient-based negotiation with expected observation},
    author = {Lanillos, Pablo and Gan, Seng Keat and Besada-Portas, Eva and Pajares, Gonzalo and Sukkarieh, Salah},
    journal = {Information Sciences},
    volume = {282},
    pages = {92-110},
    year = {2014},
    publisher = {Information Sciences}
}

@INPROCEEDINGS{ma2024mode,
  title={MoDE: CLIP Data Experts via Clustering}, 
  author={Ma, Jiawei and Huang, Po-Yao and Xie, Saining and Li, Shang-Wen and Zettlemoyer, Luke and Chang, Shih-Fu and Yih, Wen-Tau and Xu, Hu},
  booktitle={2024 IEEE/CVF Conference on Computer Vision and Pattern Recognition (CVPR)}, 
  year={2024},
}

@article{ma2024privileged,
  title={Privileged Reinforcement and Communication Learning for Distributed, Bandwidth-limited Multi-robot Exploration},
  author={Ma, Yixiao and Liang, Jingsong and Cao, Yuhong and Tan, Derek Ming Siang and Sartoretti, Guillaume},
  journal={arXiv preprint arXiv:2407.20203},
  year={2024}
}

@inproceedings{lai2024lisa,
  title={Lisa: Reasoning segmentation via large language model},
  author={Lai, Xin and Tian, Zhuotao and Chen, Yukang and Li, Yanwei and Yuan, Yuhui and Liu, Shu and Jia, Jiaya},
  booktitle={Proceedings of the IEEE/CVF Conference on Computer Vision and Pattern Recognition},
  pages={9579--9589},
  year={2024}
}

@INPROCEEDINGS{tan2024context,
      title={Context Mask Priors via Vision-Language Model for Ergodic Search}, 
      author={Tan, Derek Ming Siang and Rao, Ananya and Breitfeld, Abigail and Sartoretti, Guillaume},
      booktitle={IEEE International Conference on Robotics and Automation (Workshop on Ergodic Control)}, 
      year={2024},
      url={https://github.com/search-tta/context-mask-search-priors},
}

@article{liu2023visual,
  title={Visual instruction tuning},
  author={Liu, Haotian and Li, Chunyuan and Wu, Qingyang and Lee, Yong Jae},
  journal={Advances in neural information processing systems},
  volume={36},
  pages={34892--34916},
  year={2023}
}

@inproceedings{kirillov2023segment,
  title={Segment anything},
  author={Kirillov, Alexander and Mintun, Eric and Ravi, Nikhila and Mao, Hanzi and Rolland, Chloe and Gustafson, Laura and Xiao, Tete and Whitehead, Spencer and Berg, Alexander C and Lo, Wan-Yen and others},
  booktitle={Proceedings of the IEEE/CVF international conference on computer vision},
  pages={4015--4026},
  year={2023}
}

@inproceedings{wang2024llm,
  title={Llm-seg: Bridging image segmentation and large language model reasoning},
  author={Wang, Junchi and Ke, Lei},
  booktitle={Proceedings of the IEEE/CVF Conference on Computer Vision and Pattern Recognition},
  pages={1765--1774},
  year={2024}
}

@inproceedings{tan2024ir,
  title={Ir 2: Implicit rendezvous for robotic exploration teams under sparse intermittent connectivity},
  author={Tan, Derek Ming Siang and Ma, Yixiao and Liang, Jingsong and Chng, Yi Cheng and Cao, Yuhong and Sartoretti, Guillaume},
  booktitle={2024 IEEE/RSJ International Conference on Intelligent Robots and Systems (IROS)},
  pages={13245--13252},
  year={2024},
  organization={IEEE}
}

@inproceedings{sarkar2024visual,
  title={A visual active search framework for geospatial exploration},
  author={Sarkar, Anindya and Lanier, Michael and Alfeld, Scott and Feng, Jiarui and Garnett, Roman and Jacobs, Nathan and Vorobeychik, Yevgeniy},
  booktitle={Proceedings of the IEEE/CVF Winter Conference on Applications of Computer Vision},
  pages={8316--8325},
  year={2024}
}

@article{sarkar2023partially,
  title={A partially-supervised reinforcement learning framework for visual active search},
  author={Sarkar, Anindya and Jacobs, Nathan and Vorobeychik, Yevgeniy},
  journal={Advances in Neural Information Processing Systems},
  volume={36},
  pages={12245--12270},
  year={2023}
}

@INPROCEEDINGS{johnson2017clevr,
  title={CLEVR: A Diagnostic Dataset for Compositional Language and Elementary Visual Reasoning}, 
  author={Johnson, Justin and Hariharan, Bharath and van der Maaten, Laurens and Fei-Fei, Li and Zitnick, C. Lawrence and Girshick, Ross},
  booktitle={2017 IEEE Conference on Computer Vision and Pattern Recognition (CVPR)}, 
  year={2017},
}

@unknown{Rawte2023hallucination,
    title = {A Survey of Hallucination in "Large" Foundation Models},
    author = {Rawte, Vipula and Sheth, Amit and Das, Amitava},
    year = {2023},
}

@article{sarkar2024gomaa,
  title={Gomaa-geo: Goal modality agnostic active geo-localization},
  author={Sarkar, Anindya and Sastry, Srikumar and Pirinen, Aleksis and Zhang, Chongjie and Jacobs, Nathan and Vorobeychik, Yevgeniy},
  journal={arXiv preprint arXiv:2406.01917},
  year={2024}
}

@inproceedings{girdhar2023imagebind,
  title={Imagebind: One embedding space to bind them all},
  author={Girdhar, Rohit and El-Nouby, Alaaeldin and Liu, Zhuang and Singh, Mannat and Alwala, Kalyan Vasudev and Joulin, Armand and Misra, Ishan},
  booktitle={Proceedings of the IEEE/CVF conference on computer vision and pattern recognition},
  pages={15180--15190},
  year={2023}
}

@article{mall2023remote,
  title={Remote sensing vision-language foundation models without annotations via ground remote alignment},
  author={Mall, Utkarsh and Phoo, Cheng Perng and Liu, Meilin Kelsey and Vondrick, Carl and Hariharan, Bharath and Bala, Kavita},
  journal={arXiv preprint arXiv:2312.06960},
  year={2023}
}

@inproceedings{sastry2025taxabind,
  title={Taxabind: A unified embedding space for ecological applications},
  author={Sastry, Srikumar and Khanal, Subash and Dhakal, Aayush and Ahmad, Adeel and Jacobs, Nathan},
  booktitle={2025 IEEE/CVF Winter Conference on Applications of Computer Vision (WACV)},
  pages={1765--1774},
  year={2025},
  organization={IEEE}
}

@article{shu2022test,
  title={Test-time prompt tuning for zero-shot generalization in vision-language models},
  author={Shu, Manli and Nie, Weili and Huang, De-An and Yu, Zhiding and Goldstein, Tom and Anandkumar, Anima and Xiao, Chaowei},
  journal={Advances in Neural Information Processing Systems},
  volume={35},
  pages={14274--14289},
  year={2022}
}

@article{zhao2023test,
  title={Test-time adaptation with clip reward for zero-shot generalization in vision-language models},
  author={Zhao, Shuai and Wang, Xiaohan and Zhu, Linchao and Yang, Yi},
  journal={arXiv preprint arXiv:2305.18010},
  year={2023}
}

@inproceedings{yuan2023robust,
  title={Robust test-time adaptation in dynamic scenarios},
  author={Yuan, Longhui and Xie, Binhui and Li, Shuang},
  booktitle={Proceedings of the IEEE/CVF Conference on Computer Vision and Pattern Recognition},
  pages={15922--15932},
  year={2023}
}

@article{ren2024grounded,
  title={Grounded sam: Assembling open-world models for diverse visual tasks},
  author={Ren, Tianhe and Liu, Shilong and Zeng, Ailing and Lin, Jing and Li, Kunchang and Cao, He and Chen, Jiayu and Huang, Xinyu and Chen, Yukang and Yan, Feng and others},
  journal={arXiv preprint arXiv:2401.14159},
  year={2024}
}

@article{niroui2019deep,
  title={Deep reinforcement learning robot for search and rescue applications: Exploration in unknown cluttered environments},
  author={Niroui, Farzad and Zhang, Kaicheng and Kashino, Zendai and Nejat, Goldie},
  journal={IEEE Robotics and Automation Letters},
  volume={4},
  number={2},
  pages={610--617},
  year={2019},
  publisher={IEEE}
}

@article{van2016sentinel,
  title={Sentinel-2A MSI and Landsat 8 OLI provide data continuity for geological remote sensing},
  author={Van der Werff, Harald and Van der Meer, Freek},
  journal={Remote sensing},
  volume={8},
  number={11},
  pages={883},
  year={2016},
  publisher={MDPI}
}

@inproceedings{koreitem2020one,
  title={One-shot informed robotic visual search in the wild},
  author={Koreitem, Karim and Shkurti, Florian and Manderson, Travis and Chang, Wei-Di and Higuera, Juan Camilo Gamboa and Dudek, Gregory},
  booktitle={2020 IEEE/RSJ International Conference on Intelligent Robots and Systems (IROS)},
  pages={5800--5807},
  year={2020},
  organization={IEEE}
}

@article{lester2021power,
  title={The power of scale for parameter-efficient prompt tuning},
  author={Lester, Brian and Al-Rfou, Rami and Constant, Noah},
  journal={arXiv preprint arXiv:2104.08691},
  year={2021}
}

@article{liu2021p,
  title={P-tuning v2: Prompt tuning can be comparable to fine-tuning universally across scales and tasks},
  author={Liu, Xiao and Ji, Kaixuan and Fu, Yicheng and Tam, Weng Lam and Du, Zhengxiao and Yang, Zhilin and Tang, Jie},
  journal={arXiv preprint arXiv:2110.07602},
  year={2021}
}

@article{hurst2024gpt,
  title={Gpt-4o system card},
  author={Hurst, Aaron and Lerer, Adam and Goucher, Adam P and Perelman, Adam and Ramesh, Aditya and Clark, Aidan and Ostrow, AJ and Welihinda, Akila and Hayes, Alan and Radford, Alec and others},
  journal={arXiv preprint arXiv:2410.21276},
  year={2024}
}

@inproceedings{stevens2024bioclip,
  title={Bioclip: A vision foundation model for the tree of life},
  author={Stevens, Samuel and Wu, Jiaman and Thompson, Matthew J and Campolongo, Elizabeth G and Song, Chan Hee and Carlyn, David Edward and Dong, Li and Dahdul, Wasila M and Stewart, Charles and Berger-Wolf, Tanya and others},
  booktitle={Proceedings of the IEEE/CVF conference on computer vision and pattern recognition},
  pages={19412--19424},
  year={2024}
}

@inproceedings{wu2024clap,
  title={Large-scale Contrastive Language-Audio Pretraining with Feature Fusion and Keyword-to-Caption Augmentation}, 
  author={Yusong Wu and Ke Chen and Tianyu Zhang and Yuchen Hui and Marianna Nezhurina and Taylor Berg-Kirkpatrick and Shlomo Dubnov},
  year={2024},
  eprint={2211.06687},
  archivePrefix={arXiv},
  primaryClass={cs.SD},
}

@article{oord2018representation,
  title={Representation learning with contrastive predictive coding},
  author={Oord, Aaron van den and Li, Yazhe and Vinyals, Oriol},
  journal={arXiv preprint arXiv:1807.03748},
  year={2018}
}

@article{zhang2023patch,
  title={Patch-level contrasting without patch correspondence for accurate and dense contrastive representation learning},
  author={Zhang, Shaofeng and Zhu, Feng and Zhao, Rui and Yan, Junchi},
  journal={arXiv preprint arXiv:2306.13337},
  year={2023}
}

@article{wang2024qwen2,
  title={Qwen2-vl: Enhancing vision-language model's perception of the world at any resolution},
  author={Wang, Peng and Bai, Shuai and Tan, Sinan and Wang, Shijie and Fan, Zhihao and Bai, Jinze and Chen, Keqin and Liu, Xuejing and Wang, Jialin and Ge, Wenbin and others},
  journal={arXiv preprint arXiv:2409.12191},
  year={2024}
}

@inproceedings{he2016deep,
  title={Deep residual learning for image recognition},
  author={He, Kaiming and Zhang, Xiangyu and Ren, Shaoqing and Sun, Jian},
  booktitle={Proceedings of the IEEE conference on computer vision and pattern recognition},
  pages={770--778},
  year={2016}
}

@inproceedings{radford2021learning,
  title={Learning transferable visual models from natural language supervision},
  author={Radford, Alec and Kim, Jong Wook and Hallacy, Chris and Ramesh, Aditya and Goh, Gabriel and Agarwal, Sandhini and Sastry, Girish and Askell, Amanda and Mishkin, Pamela and Clark, Jack and others},
  booktitle={International conference on machine learning},
  pages={8748--8763},
  year={2021},
  organization={PmLR}
}

@inproceedings{chiun2025marvel,
      title={MARVEL: Multi-Agent Reinforcement Learning for constrained field-of-View multi-robot Exploration in Large-scale environments}, 
    booktitle={2025 IEEE International Conference on Robotics and Automation (ICRA)},
      author={Jimmy Chiun and Shizhe Zhang and Yizhuo Wang and Yuhong Cao and Guillaume Sartoretti},
      year={2025},
}

@inproceedings{
    hu2018overcoming,
    title={Overcoming Catastrophic Forgetting  via Model Adaptation},
    author={Wenpeng Hu and Zhou Lin and Bing Liu and Chongyang Tao and Zhengwei Tao and Jinwen Ma and Dongyan Zhao and Rui Yan},
    booktitle={International Conference on Learning Representations},
    year={2019},
}

@software{glenn2024yolo11,
  author = {Glenn Jocher and Jing Qiu},
  title = {Ultralytics YOLO11},
  version = {11.0.0},
  year = {2024},
  url = {https://github.com/ultralytics/ultralytics},
  orcid = {0000-0001-5950-6979, 0000-0002-7603-6750, 0000-0003-3783-7069},
  license = {AGPL-3.0}
}

@article{choset2001coverage,
    author = {Howie Choset},
    title = {Coverage for robotics - A survey of recent results},
    journal = {Annals of Mathematics and Artificial Intelligence},
    year = {2001},
    month = {October},
    volume = {31},
    pages = {113 - 126},
}

@misc{goyal2018accurate,
      title={Accurate, Large Minibatch SGD: Training ImageNet in 1 Hour}, 
      author={Priya Goyal and Piotr Dollár and Ross Girshick and Pieter Noordhuis and Lukasz Wesolowski and Aapo Kyrola and Andrew Tulloch and Yangqing Jia and Kaiming He},
      year={2018},
      eprint={1706.02677},
      archivePrefix={arXiv},
      primaryClass={cs.CV},
}

@misc{kaplan2020scaling,
      title={Scaling Laws for Neural Language Models}, 
      author={Jared Kaplan and Sam McCandlish and Tom Henighan and Tom B. Brown and Benjamin Chess and Rewon Child and Scott Gray and Alec Radford and Jeffrey Wu and Dario Amodei},
      year={2020},
      eprint={2001.08361},
      archivePrefix={arXiv},
      primaryClass={cs.LG},
}

@inproceedings{hu2022lora,
    title={Lo{RA}: Low-Rank Adaptation of Large Language Models},
    author={Edward J Hu and yelong shen and Phillip Wallis and Zeyuan Allen-Zhu and Yuanzhi Li and Shean Wang and Lu Wang and Weizhu Chen},
    booktitle={International Conference on Learning Representations},
    year={2022},
}

@article{ign2022flair1,
  doi = {10.13140/RG.2.2.30183.73128/1},
  author = {Garioud, Anatol and Peillet, Stéphane and Bookjans, Eva and Giordano, Sébastien and Wattrelos, Boris},
  title = {FLAIR \#1: semantic segmentation and domain adaptation dataset},
  publisher = {arXiv},
  year = {2022}
}

@misc{haarnoja2019soft,
      title={Soft Actor-Critic Algorithms and Applications}, 
      author={Tuomas Haarnoja and Aurick Zhou and Kristian Hartikainen and George Tucker and Sehoon Ha and Jie Tan and Vikash Kumar and Henry Zhu and Abhishek Gupta and Pieter Abbeel and Sergey Levine},
      year={2019},
}

@misc{simulator2022tum,
  author       = {Hongrong Huang and Jürgen Sturm},
  title        = {tum\_simulator},
  url          = {https://wiki.ros.org/tum_simulator},
}

@article{miller2016ergodic,
  author={Miller, Lauren M. and Silverman, Yonatan and MacIver, Malcolm A. and Murphey, Todd D.},
  journal={IEEE Transactions on Robotics}, 
  title={Ergodic Exploration of Distributed Information}, 
  year={2016},
  volume={32},
  number={1},
  pages={36-52},
  doi={10.1109/TRO.2015.2500441}
}

@article{vashisth2024deep,
  author={Vashisth, Apoorva and Rückin, Julius and Magistri, Federico and Stachniss, Cyrill and Popović, Marija},
  journal={IEEE Robotics and Automation Letters}, 
  title={Deep Reinforcement Learning With Dynamic Graphs for Adaptive Informative Path Planning}, 
  year={2024},
  volume={9},
  number={9},
  pages={7747-7754},
  doi={10.1109/LRA.2024.3421188}
}

@article{moon2022tigris,
  author={Moon, Brady and Chatterjee, Satrajit and Scherer, Sebastian},
  booktitle={2022 IEEE/RSJ International Conference on Intelligent Robots and Systems (IROS)}, 
  title={TIGRIS: An Informed Sampling-based Algorithm for Informative Path Planning}, 
  year={2022},
  volume={},
  number={},
  pages={5760-5766},
  doi={10.1109/IROS47612.2022.9981992}
}

\newpage

\appendix

\begin{center}
    \Large\textbf{Appendix: Supplementary Material}
\end{center}

\captionsetup{font=normalsize}

\renewcommand{\thetable}{\Alph{section}.\arabic{table}}
\renewcommand{\thefigure}{\Alph{section}.\arabic{figure}}
\renewcommand{\theequation}{\Alph{section}.\arabic{equation}}
\counterwithin{table}{section}
\counterwithin{figure}{section}
\counterwithin{equation}{section}

\section{AVS-Bench Dataset Details}
\label{app:dataset}

\begin{figure}[htbp]
    \vspace{2mm}
    \centering
    \includegraphics[width=1.0\linewidth]{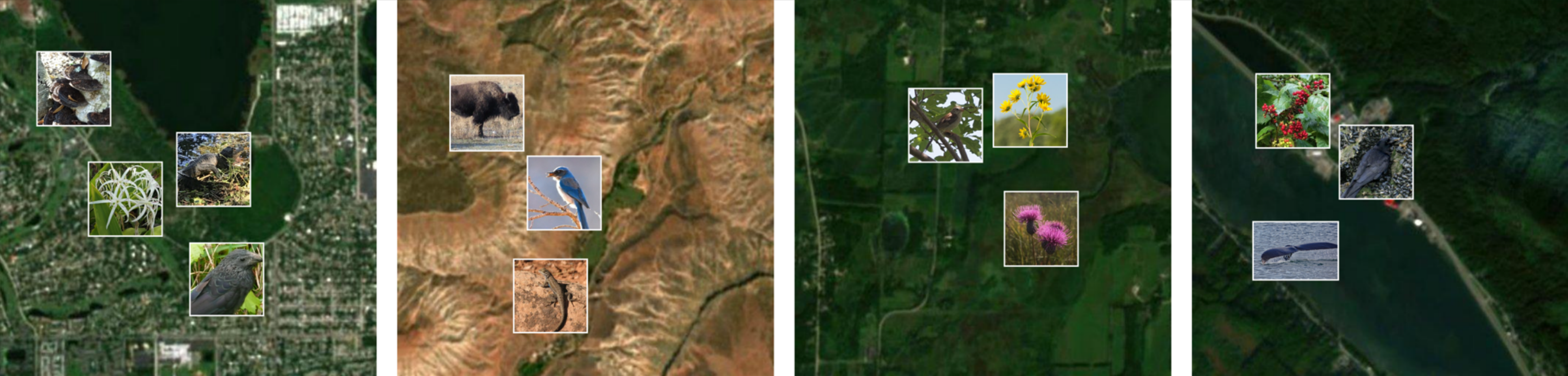}
    \vspace{-0.5cm}
    \caption{Examples of satellite images~\cite{van2016sentinel} in the full \textbf{380k} dataset (each with different taxonomies~\cite{van2021inat}), used for CLIP fine-tuning.}
    \label{fig:full-ds-examples}
\end{figure}

\vspace{5mm}
\begin{figure}[htbp]
    \centering
    \includegraphics[width=1.0\linewidth]{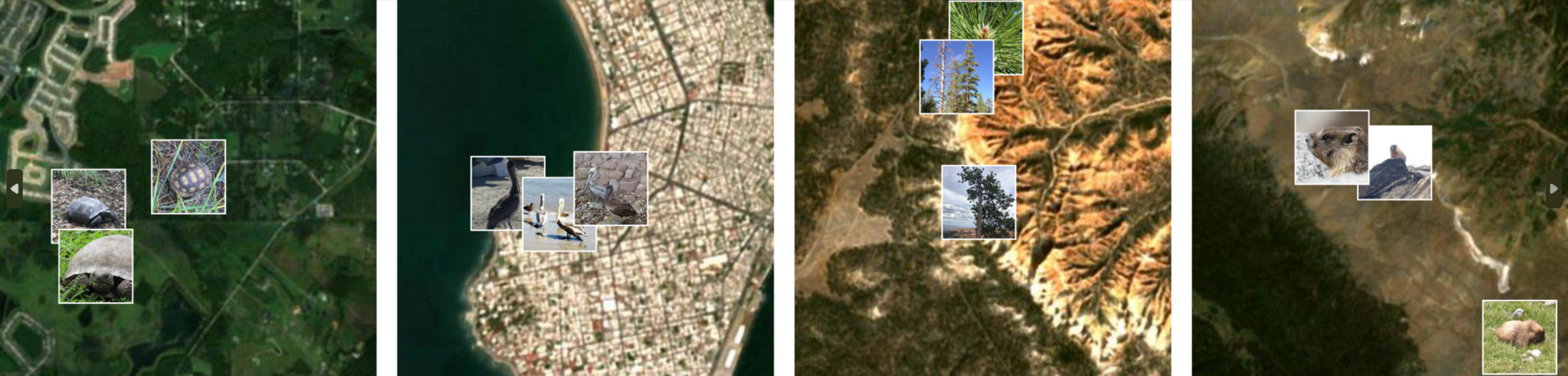}
    \vspace{-0.5cm}
    \caption{Examples of satellite images~\cite{van2016sentinel} in the \textbf{80k} training and \textbf{4k} validation datasets (each with same taxonomies~\cite{van2021inat}), used for AVS validation and score maps generation.}
    \label{fig:avs-ds-examples}
\end{figure}
\vspace{3mm}

In this section, we provide more details about our AVS-Bench dataset composition and generation process, in addition to the information provided in Sec.~\ref{sec:dataset-generation}.
Our \textit{tri-modal} dataset split contains Sentinel-2 level 2A satellite images~\cite{van2016sentinel} covers approximately 2.56km$\times$2.56km of land mass, each with targets that are paired with their taxonomic names, locations, and ground images. 
\vspace{-2.0mm}
\begin{enumerate}
    \item \textbf{CLIP training dataset}: 380k satellite images with different taxonomic targets (Fig.~\ref{fig:full-ds-examples}).
    \vspace{-1.5mm}
    \item \textbf{AVS training dataset}: 80k satellite images with same taxonomic targets (Fig.~\ref{fig:avs-ds-examples}).
    \vspace{-1.5mm}
    \item \textbf{AVS validation datasets}: 4k satellite images with same taxonomic targets that are \textit{in-domain}, and 4k satellite images with same taxonomic targets that are \textit{out-domain} (Fig.~\ref{fig:avs-ds-examples}).   
\end{enumerate}

\subsection{Geographical Coverage}
\label{app:dataset:coverage}

\begin{figure}[h]
    \vspace{0.2cm}
    \centering
    \begin{minipage}[t]{0.495\textwidth}
        \centering
        \includegraphics[width=\linewidth]{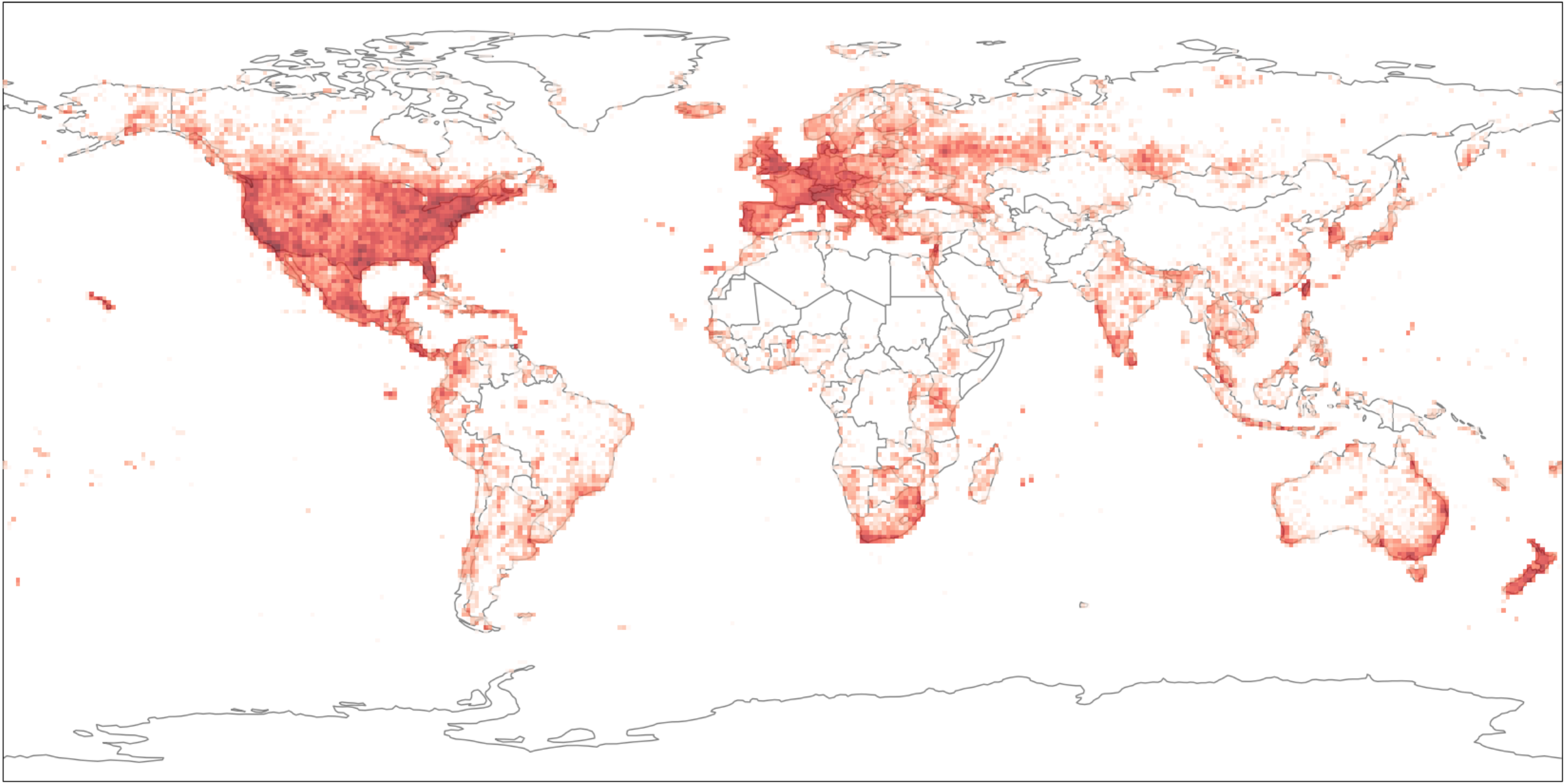}
        \par\smallskip
        \centering \textbf{380k} full training dataset 
    \end{minipage}
    \hfill
    \begin{minipage}[t]{0.495\textwidth}
        \centering
        \includegraphics[width=\linewidth]{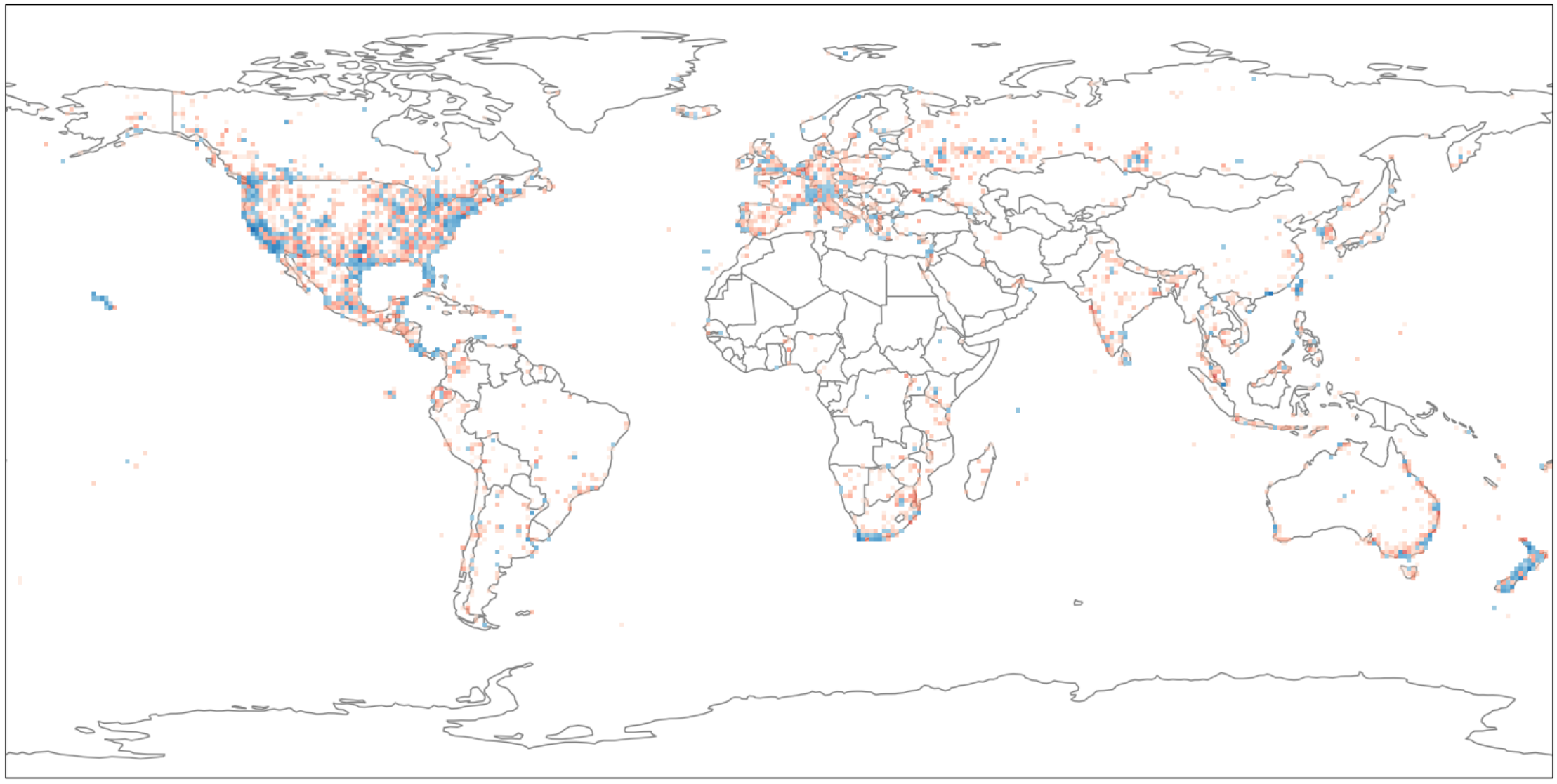}
        \par\smallskip
        \centering \textbf{80k} train (Red) \& \textbf{4k} validation (Blue) datasets
    \end{minipage}

    \caption{Geographic coverage of datasets used in training and validation.}
    \label{fig:spatial-distribution-tax}
\end{figure}

We visualize the spatial distribution of our dataset in Fig.~\ref{fig:spatial-distribution-tax}, where the color intensity reflects the taxonomy counts in each cell ($\text{1}^\circ$ latitude $\times$ $\text{1}^\circ$ longitude).
Despite filtering our dataset to cater to our AVS task, the \textbf{80k} training and \textbf{4k} \textit{in-domain} validation datasets appear visually representative of the original \textbf{380k} dataset distribution (\textbf{4k} \textit{out-domain} validation dataset has a similar distribution). 

\subsection{Taxonomy Statistics}
\label{app:dataset:stats}

\begin{figure}[t] 
    \centering
    \includegraphics[width=\linewidth]{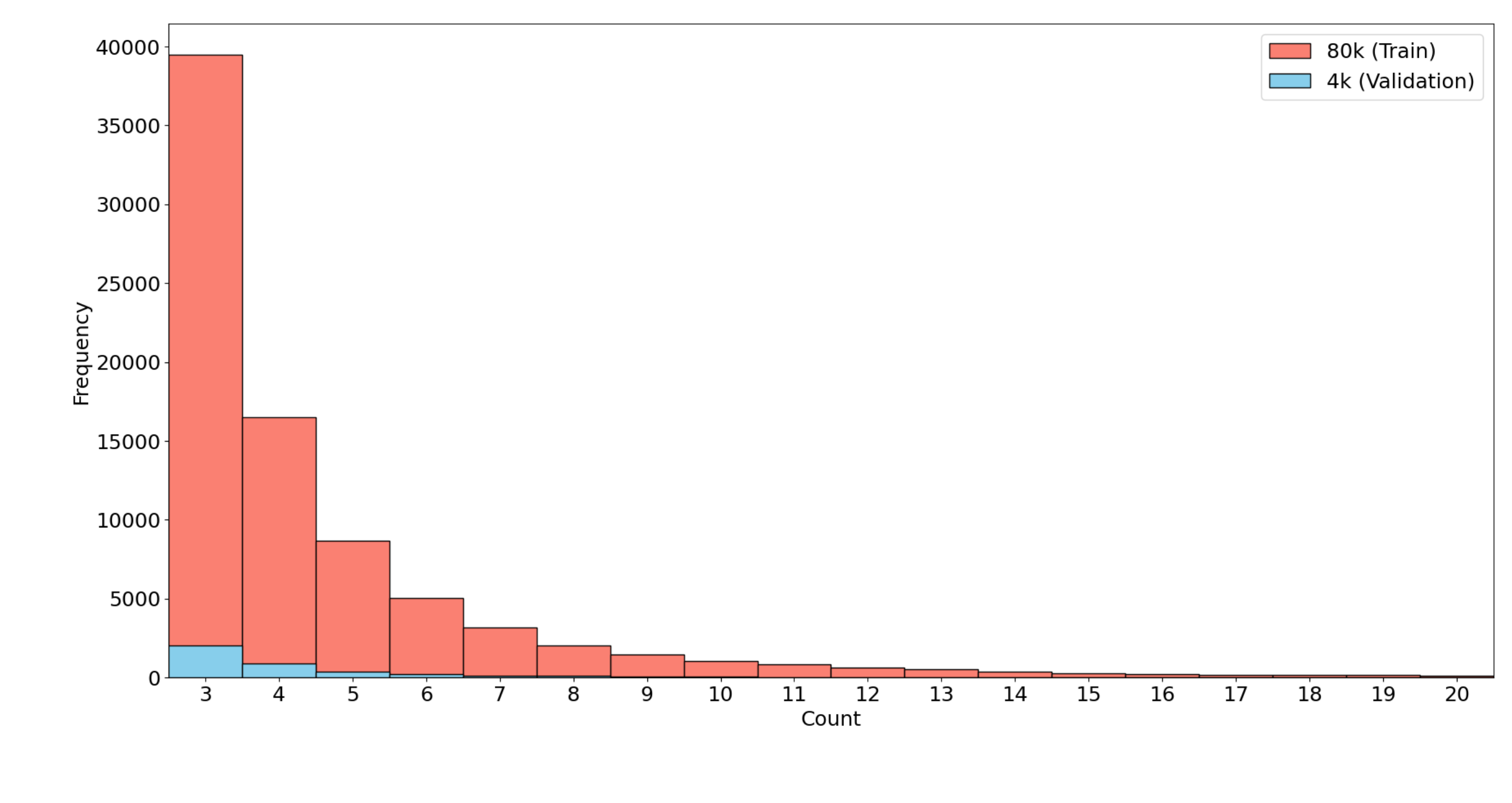}
    \vspace{-1.0cm}
    \caption{Histogram counts for \textbf{80k} train (Red) \& \textbf{4k} \textit{out-domain} validation (Blue) datasets}
    \label{fig:tax-counts-hist}
\end{figure}

We visualize the distribution of the taxonomy counts for the training and validation datasets in Fig.~\ref{fig:tax-counts-hist}.
We note a natural decreasing trend in the number of same-taxonomy targets within the 2.56km$\times$2.56km of land mass covered by each image.
The average number of taxonomic counts per image for our \textbf{80k} training datset is 4.5$\pm$2.6, \textbf{4k} \textit{in-domain} validation dataset is 3.3$\pm$0.7, and \textbf{4k} \textit{out-domain} validation dataset is 4.3$\pm$2.3.
This results in a very sparse target distribution per image, making our AVS task challenging given a budget constraint.

In addition, we visualized the breakdown of the taxonomy categories for all of our datasets in Table ~\ref{tab:dataset_distribution}.
These distributions, except for the \textit{out-domain} validation dataset, are similar to the dataset distribution from the iNaturalist 2021 challenge where these datasets originate from~\cite{van2021inat}.  

\begin{table}[t]
    \vspace{-0.3cm}
    \centering
    \scriptsize
    \captionof{table}{Taxonomy distribution across training and validation datasets}
    \setlength{\tabcolsep}{0pt} 
    \begin{tabularx}{\linewidth}{
        >{\raggedright\arraybackslash\hspace{0pt}}X 
        @{\hspace{2mm}}
        >{\centering\arraybackslash}X
        >{\centering\arraybackslash}X
        >{\centering\arraybackslash}X
        >{\centering\arraybackslash}X
        >{\centering\arraybackslash}X
        >{\centering\arraybackslash}X
        >{\centering\arraybackslash}X
        >{\centering\arraybackslash}X}
    \toprule
    \multirow{2}{*}{\textbf{Taxonomies}}
    & \multicolumn{2}{c}{\textbf{Train (380k)}}
    & \multicolumn{2}{c}{\textbf{Train (80k)}}
    & \multicolumn{2}{c}{\textbf{Val In-Domain (4k)}}
    & \multicolumn{2}{c}{\textbf{Val Out-Domain (4k)}} \\
    \cmidrule(lr){2-3} \cmidrule(lr){4-5} \cmidrule(lr){6-7} \cmidrule(lr){8-9}
    & Targets (\%) & Images (\%) 
    & Targets (\%) & Images (\%)
    & Targets (\%) & Images (\%)
    & Targets (\%) & Images (\%) \\
    \midrule
    Plants & 43.8 & -- & 45.3 & 45.8 & 51.1 & 50.9 & 39.7 & 40.1 \\
    Insects & 30.1 & -- & 29.8 & 29.5 & 16.9 & 17.1 & 1.1 & 1.2 \\
    Birds & 12.7 & -- & 11.5 & 11.5 & 17.4 & 17.3 & 18.7 & 19.8 \\
    Fungi & 3.5 & -- & 2.7 & 2.9 & 2.5 & 2.6 & 1.9 & 2.1 \\
    Reptiles & 3.8 & -- & 3.5 & 3.4 & 4.2 & 4.2 & 0.7 & 0.6 \\
    Mammals & 0.8 & -- & 1.1 & 1.1 & 0.9 & 1.0 & 8.4 & 8.3 \\
    Fishes & 1.0 & -- & 1.7 & 1.5 & 1.8 & 1.9 & 3.7 & 3.7 \\
    Amphibians & 2.1 & -- & 2.4 & 2.3 & 1.8 & 1.9 & 0.0 & 0.0 \\
    Mollusks & 0.2 & -- & 0.5 & 0.4 & 0.6 & 0.6 & 14.9 & 14.2 \\
    Arachnids & 1.6 & -- & 1.0 & 1.1 & 1.1 & 1.1 & 0.5 & 0.5 \\
    Animalia & 0.4 & -- & 0.5 & 0.5 & 1.6 & 1.7 & 10.3 & 9.6 \\
    \midrule
    Total & 2,601,787 & 379,962 & 365,292 & 80,535 & 13,334 & 4000 & 17,188 & 4000 \\
    \bottomrule
    \end{tabularx}
    \label{tab:dataset_distribution}
\end{table}

\subsection{Score Map Generation}
\label{app:dataset:score-map-gen}

We generate our \textbf{80k} training score masks using a custom process because our AVS dataset only includes point locations, and conversion to segmentation masks with likelihood scores is non-trivial.
This is done in two stages.
First, we use GSNet~\cite{ye2025GSNet}, an open-vocabulary semantic segmentation model, to obtain label maps of the low-resolution Sentinel-2 images based on broad landmark names (i.e. \textit{Urban}, \textit{Water}, \textit{Vegetation}, \textit{Barren}). 
Since GSNet has been pretrained on a diverse set of satellite images with varying spatial resolution, we fine-tune it with the FLAIR semantic segmentation dataset~\cite{ign2022flair1} to enhance its segmentation abilities specifically for low-resolution images.
However, despite the high-quality segmentation masks, mislabeling commonly occurs. 
Hence, we use GPT4o~\cite{hurst2024gpt}, augmented with human-labeled examples, to rectify the wrong labels.
We then use GPT4o again to score the masks based on the likelihood of finding the specified taxonomies per image.
It concurrently generates a conversation that explains the rationale for why certain landmarks are more suitable for the specified taxonomies, and less so for others. 
After filtering erroneous GPT4o generations, we end up with \textbf{80k} score maps for each taxonomic-image pair. 
An example can be seen in Fig.~\ref{fig:score-map-viz} (prompts used for relabeling and scoring by GPT4o are shown in Appendix~\ref{app:prompt:gpt4o-score-maps}).

\begin{figure}[t] 
    \centering
    \includegraphics[width=1.0\linewidth]{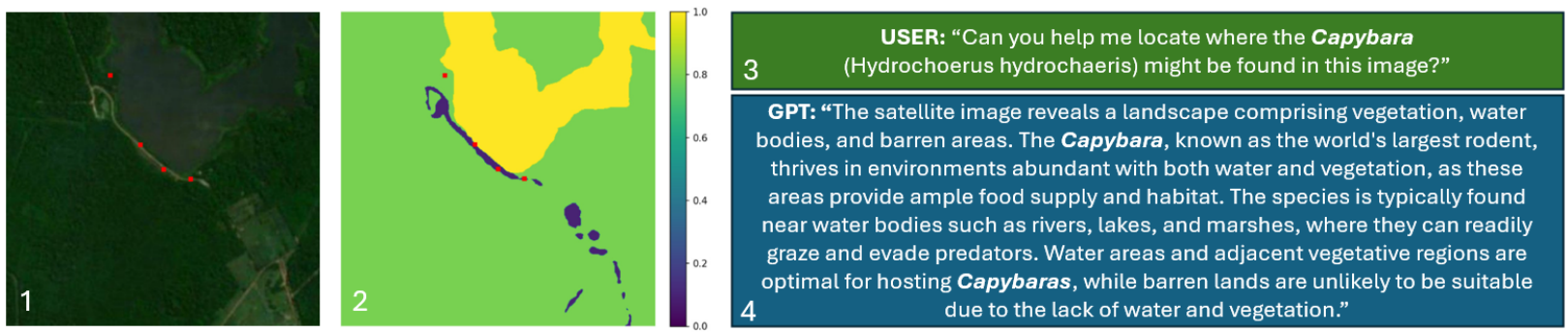}
    \caption{(1) Satellite map~\cite{van2016sentinel} where Capybaras can be found~\cite{van2021inat}. (2) Score map of where Capybaras are likely to be found, used for fine-tuning VLM baselines and training RL policy. (3-4) Question and answer pair, used to fine-tune VLMs such as LISA~\cite{lai2024lisa}.}
    \label{fig:score-map-viz}
\end{figure}

\begin{figure} [t]
    \centering
    \includegraphics[width=1.0\linewidth]{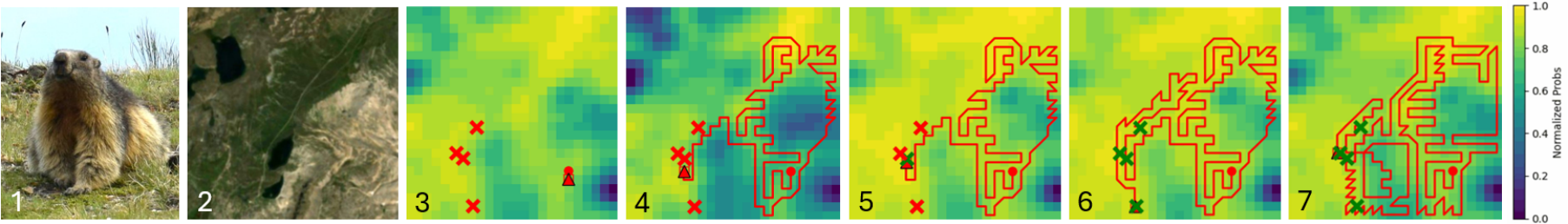}
    \caption{(1) Ground image of a \textit{Marmot}. (2) Satellite image where marmots can be found. (3) Initial probability output. (4) TTA enables regulated decrease in probability values in regions where marmots are not detected. (5) TTA significantly increases probability values in region where the first Marmot is found. (6) Improved priors leads to efficient search. (7) Inefficient search without TTA. }
    \label{fig:tta-comparison-viz}
\end{figure}

\subsection{Sound Dataset Generation}
\label{app:dataset:sound-dataset-gen}

We further fine-tune a sound encoder in order to evaluate the generalization ability of Search-TTA to previously unseen input modalities (Sec.~\ref{sec:expt-multimodal-inputs}).
We follow a similar process as in Sec.~\ref{sec:dataset-generation} to generate our \textit{quad-modal} training and validation datasets.
We begin with the iSoundNat dataset~\cite{sastry2025taxabind} with \textbf{74.9k} satellite images, each matching a ground-level image, a taxonomic label, and a sound recording.
We keep only the \textit{in-domain} taxonomies defined in Sec.~\ref{sec:dataset-generation}, and end up with \textbf{68.8k} data points used for fine-tuning the CLAP sound encoder~\cite{wu2024clap}. 
To validate the its performance, we curate a new validation dataset by keeping data from the \textbf{4k} in-domain validation datasets that contain targets with sound data.  
By the end of this process, we have an in-domain validation dataset with \textbf{480} data points containing all modalities including sound.  
Training details can be found in Appendix~\ref{app:search-tta:training-sat-encoder}.

\section{Additional Search-TTA Details}
\label{app:search-tta}

In this section, we provide more details of the Search-TTA framework, to supplement the information in Sec.~\ref{sec:search-tta-framework}.
We elaborate on how Search-TTA generates and periodically updates the probability distribution outputs that serve as a strong prior to the RL search planner.

\subsection{Qualitative Analysis}
\label{app:search-tta:qualitative}

We provide snapshots of AVS with the RL planner to provide a better understanding of how it works.
In Fig.~\ref{fig:tta-comparison-viz}, the RL planner begins with an initial prior generated by the satellite image CLIP encoder, which represents the probability distribution of where \textit{Marmots} (\textit{Mammalia Rodentia Sciuridae Marmota}) can be found. 
Within the first phase, TTA enables a regulated decrease in probability values within the regions where marmots are not detected. 
Thereafter, Search-TTA significantly improves the probability distribution outputs in the associated region where the first and subsequent targets are found. 
This steers the RL planner to exploit the high probability region to locate all of the targets within 181 steps.
Without TTA, the probability distribution is static and the RL planner takes 242 steps before locating all marmots.

\subsection{Algorithm}
\label{app:search-tta:algorithm}
\vspace{-1mm}
We provide the pseudo-code to illustrate the Search-TTA framework (algorithm~\ref{alg:search-tta-algo}).

\begin{algorithm}[]
\caption{The Search-TTA Framework (with RL Search Policy)}
\label{alg:search-tta-algo}
{\small
\SetKwInOut{Input}{Input}
\Input{
  Satellite-image encoder \(f_{\theta}\); Query modality encoder \(h_{\psi}\); Search policy \(\pi_{\phi}\), \\
  Satellite image \(s\), Query modality \(g\), Budget \(\mathcal{B}\)
}
\textbf{Initialize:} \(\theta \gets \theta_{\rm base}\), step \(t=0\), measurements \(O=[0, \dots, 0]\), grid map \(\mathcal{M} = (n \times n)\), \(\alpha_{\text{pos},i} = 4\)    \\

\tcp{--- Generate Probability Distribution ---}
\(z \gets f_{\theta}(s)\), \(y \gets h_{\psi}(g)\)\;
\(P \gets \mathrm{cosineSim}(z,y)\)\;
\(r \gets \mathrm{kmeans}(z)\)\;
\While{\(t \leq \mathcal{B}\)}{
  \tcp{--- Generate Action ---}
  \(a_t \sim \pi_\phi(\cdot\mid o_t, P)\)\;
  \(s_{t+1} \sim T(o_t, a_t)\)\;
  \tcp{--- Collect Measurements ---}
  \(d \gets \mathrm{Observe}(o_{t+1})\)\;
  \(O \gets \mathrm{Update}(O, d)\)\;
  
  \tcp{--- Perform TTA ---}
  \For{every \(k\) steps}{
    \(\theta\leftarrow\theta_{\rm base}\)\;
    \(\gamma \gets \gamma_{\rm min} + (t / n^2) \cdot (\gamma_{\rm max} - \gamma_{\rm min})\)\;
    \(\alpha_{\text{neg},j} \gets \min \left(\beta \left ( O_r / L_r \right)^{\gamma},\ 1 \right)\)\; 
    \(L(\lambda) = \sum_{i=1}^{p} \alpha_{\text{pos},i} \cdot \log \lambda(x_i) - \sum_{j=1}^{n} \alpha_{\text{neg},j} \cdot \lambda(x_j) \, dx\), where \(P \approx \lambda\)\;
    \(\theta \leftarrow \theta + \gamma \cdot \nabla_{\theta}L(\lambda)\)\;
  }
}
}
\end{algorithm}

\subsection{ Training Details for Satellite Image and Sound Encoders}
\label{app:search-tta:training-sat-encoder}
\vspace{-1mm}

We fine-tune our satellite image CLIP encoder~\cite{radford2021learning} with the hyperparameters in Table~\ref{tab:taxabind_training}. 
This was performed using two NVIDIA A6000 GPUs, which took 3 epochs (3.5 days) before convergence (lowest CLIP validation score).
In addition, we fine-tune our CLAP sound encoder~\cite{wu2024clap} with similar hyperparameters.
This was performed using four NVIDIA A5000 GPUs, which took 19 epochs (11 hours).
While fine-tuning both encoders, we keep our BioCLIP~\cite{stevens2024bioclip} model frozen.

\begin{table}[htbp]
    \vspace{-0.1cm}
    \small
    \centering
    \caption{Training hyperparameters (query encoder)}
    \begin{tabular}{ll}
        \toprule
        \textbf{Hyperparameter} & \textbf{Value} \\
        \midrule
        Batch Size & 32 \\
        Learning Rate & 1e-4 \\
        Learning Rate Schedule & min 1e-6 (Cosine Annealing) \\
        Temperature ($\tau$) & 0.07 \\
        Optimizer & AdamW \\
        Optimizer $\beta$ & (0.9, 0.98) \\
        Optimizer $\epsilon$ & 1e-6 \\
        Accumulate Grad batches & 64 \\
        Projection Dimension & 512 \\
        Satellite Image Encoder & CLIP~\cite{radford2021learning} (ViT-L/14@336px) \\
        Ground Image Encoder & BioCLIP~\cite{stevens2024bioclip} (ViT-B/16) \\
        Text Encoder & BioCLIP~\cite{stevens2024bioclip} \\
        Sound Encoder & CLAP~\cite{wu2024clap} \\
        \bottomrule
    \end{tabular}
    \vspace{0.5em}
    \vspace{-0.1cm}
    \label{tab:taxabind_training}
\end{table}

\subsection{Kmeans Clustering}
\label{app:search-tta:kmeans}

We rely on Kmeans clustering of the satellite image encoder output to obtain clusters of embeddings that are deemed semantically similar by CLIP~\cite{ma2024mode}.
We determine the best $k$ by taking the average of the silhouette score criterion~\cite{shahapure2020silhouette} and the elbow criterion~\cite{marutho2018elbow}. 
The silhouette score measures clustering quality by contrasting each satellite patch feature's average distance to its own cluster with the closest alternative cluster.
On the other hand, the elbow method charts the within‑cluster sum‑of‑squares across candidate $k$ values and pinpoints where additional clusters yield only marginal variance reduction.
Combining both methods balances the silhouette score's tendency to favor fewer clusters with the elbow method's subjectivity based on a 'knee' point.
In practice, we set the max averaged $k$ to be 4 based on the approximate number of possible broad semantic landmarks, should the elbow method overestimate $k$ too significantly.

\subsection{Varying SPPP-based Online Adaptation Hyperparameters}
\label{app:search-tta:sppp}

We perform grid search to determine the optimal hyperparameters for our negative weighting coefficient $\alpha_{\text{neg,j}} =  \min \left(\beta \left ( O_r / L_r \right)^{\gamma},\ 1 \right)$, where $O_r$ is the number of patches observed in region $r$ and $L_r$ is the number of patches in that region.
$\beta$ balances the relative weightage between positive and negative measurements in the loss function, while $\gamma$ scales the weightage of negative measurements given the same amount of region explored. 
We summarize our results in Table~\ref{tab:expt-param-grid-search}.
When we remove the negative weighting coefficient ($\gamma=$ 0), we observe poor performance. 
This is because all negative samples are weighted equally heavily even at the start of AVS, causing premature collapsing of probability distribution modes. 
Hence, this highlights the importance of our uncertainty weighting scheme.
In addition, if we remove the relative weighting factor ($\beta=$ 1), we note one of the worst performances possibly due to over-penalizing negative measurements.

In order to achieve stable updates to the output probability distribution, we reset the satellite encoder weights back to the base weights before running TTA updates.
During TTA updates, we use the Adam optimizer, and employ a learning rate schedule that increases our learning rate from 1e-6 to 1e-5 depending on how much of the search space has been covered. 
This learning rate schedule allows the model to learn more effectively when more measurements are collected~\cite{goyal2018accurate}.

\begin{table}[t]
\vspace{-0mm} \tiny
\centering
\caption{Effect of SPPP weighting coefficient on targets found (\%) $\uparrow$}
\resizebox{\columnwidth}{!}{
\setlength{\tabcolsep}{3pt}
\begin{tabular}{cc ccc ccc @{}p{0.6em}@{} ccc ccc}
\toprule
\multicolumn{2}{c}{}
    & \multicolumn{6}{>{\columncolor{headergreen}}c}{\textbf{In-domain}}
    & 
    & \multicolumn{6}{>{\columncolor{headergreen}}c}{\textbf{Out-domain}} \\

\addlinespace[1mm]
\multicolumn{2}{l}{\raisebox{0.5\normalbaselineskip}[0pt][0pt]{\textbf{Parameters}}}
    & \multicolumn{3}{c}{\textbf{$\mathcal{B}=256$}}
    & \multicolumn{3}{c}{\textbf{$\mathcal{B}=384$}}
    &
    & \multicolumn{3}{c}{\textbf{$\mathcal{B}=256$}}
    & \multicolumn{3}{c}{\textbf{$\mathcal{B}=384$}} \\
\cmidrule(lr){3-5} \cmidrule(lr){6-8} \cmidrule(lr){10-12} \cmidrule(lr){13-15}
\noalign{\vskip -0.3ex}
\textbf{$\beta$} & \textbf{$\gamma$}
    & {\tiny\textbf{\textit{All}}} & {\tiny\textbf{\textit{Bot. 5\%}}} & {\tiny\textbf{\textit{Bot. 2\%}}}
    & {\tiny\textbf{\textit{All}}} & {\tiny\textbf{\textit{Bot. 5\%}}} & {\tiny\textbf{\textit{Bot. 2\%}}}
    &
    & {\tiny\textbf{\textit{All}}} & {\tiny\textbf{\textit{Bot. 5\%}}} & {\tiny\textbf{\textit{Bot. 2\%}}}
    & {\tiny\textbf{\textit{All}}} & {\tiny\textbf{\textit{Bot. 5\%}}} & {\tiny\textbf{\textit{Bot. 2\%}}} \\
\noalign{\vskip -0.4ex} 
\midrule
\sfrac{1}{9} & 2 & \textbf{57.4} & \textbf{28.0} & 27.3 & \textbf{76.1} & \textbf{53.0} & \textbf{51.9} & & \textbf{60.8} & \textbf{31.7} & \textbf{30.7} & \textbf{79.6} & \textbf{58.9} & \textbf{56.1} \\
\sfrac{1}{9} & 1 & 57.0 & 27.6 & \textbf{27.9} & 75.2 & 51.3 & 51.9 & & 60.3 & 29.6 & 25.7 & 78.8 & 54.9 & 51.4 \\
\sfrac{1}{9} & 0 & 57.1 & 26.9 & 23.9 & 75.7 & 50.3 & 48.8 & & 59.7 & 27.7 & 22.6 & 78.6 & 53.5 & 50.2 \\
1            & 2 & 56.3 & 27.0 & 25.8 & 75.0 & 50.5 & 47.9 & & 60.0 & 30.2 & 27.5 & 78.4 & 55.9 & 51.8 \\
\midrule
-- & --      & 56.6 & 26.6 & 27.3 & 75.5 & 49.9 & 51.4 & 
             & 58.5 & 23.1 & 16.0 & 77.1 & 44.8 & 36.1 \\
\bottomrule
\end{tabular}
}
\label{tab:expt-param-grid-search}
\vspace{-4mm}
\end{table}

\section{Additional Baseline Details}
\label{app:baselines}

In this section, we provide additional information on how we set up our baselines for fair comparison, on top of the details provided in Sec.~\ref{sec:experiments}.

\subsection{Planner Baselines}
\label{app:baselines:planner}

We compare Search-TTA with an Attention-based Reinforcement Learning (RL) planner~\cite{cao2023ariadne} and a greedy Information Surfing (IS) planner~\cite{lanillos2014multi}.
The RL planner is non-myopic in nature as it learns dependencies at multiple spatial scales across the entire search domain.
This allows agents to balance trade-offs between short-term exploitation and long-term exploration given the probability distribution map.
On the other hand, the IS planner drives agents in the direction of the derivative of the information map to maximize short-term gains.
Such an approach tends to be greedy in nature and may suffer from overexploitation of local maxima.
By design, the RL planner allows movement to all eight neighboring cells, while the IS planner is limited to the four cardinal directions.
Note that we do not intend to compare the performance between RL and IS, but rather how test-time adaptation improves each of the planners independently. 
Lastly, we use a lawnmower planner~\cite{choset2001coverage} as a weak baseline for comparison. 
Starting from the top-left grid, the lawnmower planner moves in a zigzag manner across each row before moving down to the next row when the current row is completely covered. 
This approach, while not using any probability distribution to direct the coverage process, provides an upper bound to the number of cells that can be covered given a specified budget. 

\begin{figure}[t] 
    \centering
    \includegraphics[width=1.0\linewidth]{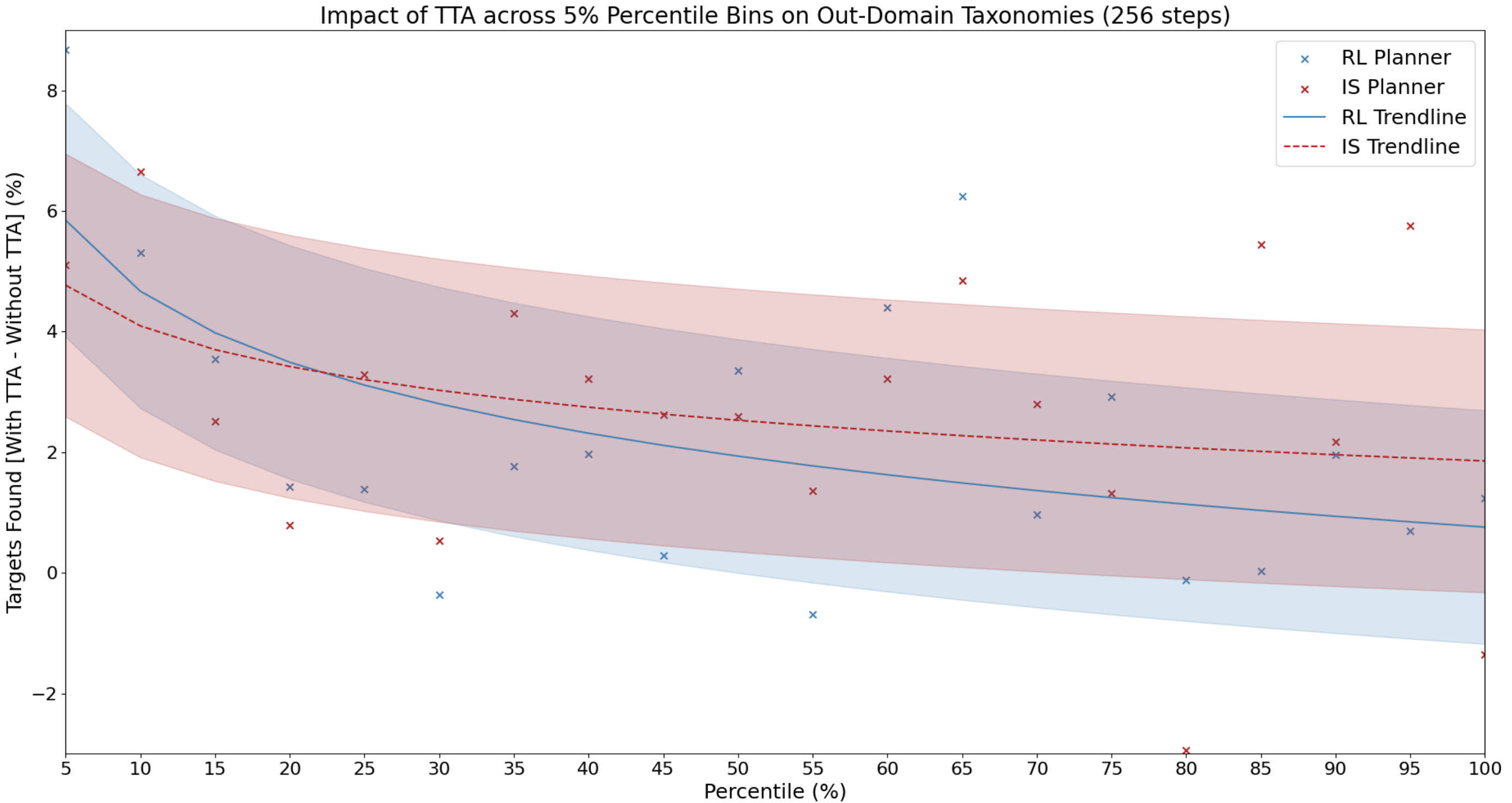}
    \caption{Performance difference (due to TTA) for RL (blue) and IS (red) planners at 256 steps.}
    \label{fig:tta-diff-trendline-256}
\end{figure}

\begin{figure}[t] 
    \centering
    \includegraphics[width=1.0\linewidth]{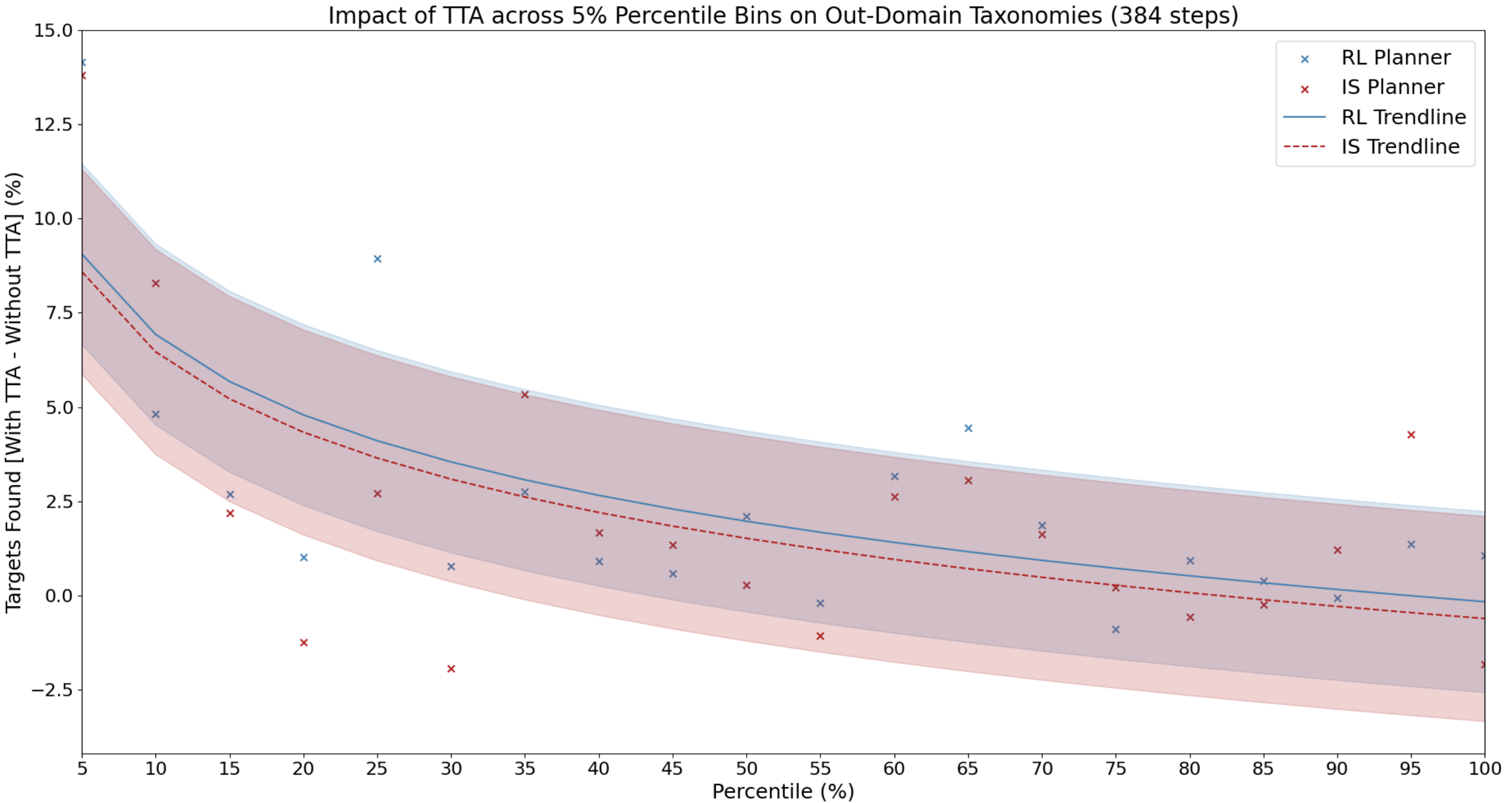}
    \caption{Performance difference (due to TTA) for RL (blue) and IS (red) planners at 384 steps.}
    \label{fig:tta-diff-trendline-384}
    \vspace{-0.6cm}
\end{figure}

\textbf{RL Policy Training:}
We train our RL planner's attention-based neural network using the soft actor-critic (SAC) algorithm~\cite{haarnoja2019soft}, which learns a policy by maximizing return while maximizing entropy.
\begin{equation}
    \pi^*= \mathop{\rm {argmax}} \mathbb{E}_{(o_t, a_t)}[\sum^T_{t=0}\gamma^t(r_t+\alpha \mathcal{H}(\pi(.|o_t)))],
\end{equation}
where $\mathcal{H}$ denotes entropy, $\pi^*$ the optimal policy, $\gamma$ the discount factor, and $\alpha$ the adaptive temperature term. 
We utilize a subset of the score maps of varying probability distributions from Sec.~\ref{sec:dataset-generation} to pre-train our RL policy.
Similar to~\cite{cao2023ariadne, tan2024ir}, we define the viewpoints $\psi$ in the search domain $\mathcal{M}$ as graph vertices, each connected via edges to its adjacent ($\leq$ 8) neighbors.
In addition to positional information, each of these nodes are augmented with the agent's visitation history and scores from Search-TTA's output probability distribution.
This graph can then be used as the RL agent's observation to output a stochastic policy $\pi_{\theta}(a^t|o^t)$.
Our reward function penalizes distance travelled while incentivizes agents to to travel to high probability regions. 
We find that adding rewards for targets found causes the planner to generate inefficient routes, possibly due to the sparse target distributions that may provide confusing reward signals.
We train our policy using an AMD Ryzen threadripper 3970x and four NVIDIA A5000 GPUs, which took 20k episodes (80 hours) to converge.

\textbf{Bottom Percentile Comparison:}
To determine Search-TTA's effectiveness given poor CLIP predictions, we break
down the percentage of targets found into the bottom 5\% and 2\% percentiles.
We measure the quality of CLIP predictions by taking the average scores of the pixels where the targets are located on the predicted score map, and deem a CLIP prediction to be poor if most targets are located in the lowest-scoring regions.
We then plot the performance gains (due to TTA) for both RL and IS planners for both 256 (Fig.~\ref{fig:tta-diff-trendline-256}) and 384 steps (Fig.~\ref{fig:tta-diff-trendline-384}), given bins of 5-percentile increments. Note that the logarithmic trendlines fit the datapoints well, indicating that the performance difference is most significant at the bottom percentiles. This illustrates how Search-TTA is particularly effective given poor score maps in the low percentile range.

\subsection{VLM Baselines}
\label{app:baselines:vlm}

We evaluate the effectiveness of Search-TTA's CLIP vision backbone by replacing it with different state-of-the-art VLMs. 
Most of these VLMs comprise a reasoning module (e.g. LLaVA~\cite{liu2023visual} or Qwen2-VL~\cite{wang2024qwen2}) that processes image and text inputs to output reasoning embeddings. 
These embeddings are then passed into segmentation modules such as SAM~\cite{kirillov2023segment} to generate the appropriate masks.
LISA~\cite{lai2024lisa} (we use LISA-7B) connects and fine-tunes LLaVA and SAM modules in an end-to-end manner. 
Unlike the original setup where LISA is trained on binary masks, we directly fine-tune LISA with the \textbf{80k} score maps (with text-based question-answer) from our custom training dataset.
In addition, we remove the final threshold layer in the SAM module to output continuous score distributions. 
On the other hand, LLM-Seg~\cite{wang2024llm} (which uses LLaVA-7B) decouples the modules by initially assigning SAM the task of producing mask proposals, which are then evaluated and chosen by LLaVA.
Unlike LISA, since there are no straightforward methods to fine-tune LLM-Seg with continuous score maps directly, we apply a binary threshold to our score maps for training (without text-based question-answer).
We also incorporate the scores generated by LLM-Seg into all binary mask proposals to output a score map.  
Lastly, we introduce two fully decoupled baselines that output landmark names and scores from LLaVA-13B and Qwen2-VL-7B.
These landmark names are then fed as input into GroundedSAM~\cite{ren2024grounded} to generate segmentation masks, which are then aggregated with the VLM's per-region scores to obtain score masks. 
All of these score maps are passed into the RL policy for path planning~\cite{tan2024context}.
Note that unlike LISA and LLM-Seg, we do not fine-tune LLaVA and Qwen2-VL.
The prompts used are shown in Appendix~\ref{app:prompt:vlm-baselines}.

\textbf{Scaling Up VLM Training Dataset: }
To justify the score map dataset size of \textbf{80k}, we experiment with varying the amount of training data used to fine-tune our strongest vision model baseline (LISA-7B) . 
As seen from Table~\ref{tab:expt-vary-vlm-train-ds}, scaling up the dataset generally improves search performance. 
However, the performance gain becomes more marginal when we scale our dataset from 55k to 80k. 
This indicates generating additional data may not yield significant performance gain.
Hence, it is a reasonable choice to stop at a dataset size of \textbf{80k} (also due to the cost of running GPT4o).

\begin{table}[t]   \scriptsize
    \centering
    \scriptsize
    \caption{Scaling up dataset size for fine-tuning LISA VLM~\cite{lai2024lisa} - targets found (\%) $\uparrow$}  
    \setlength{\tabcolsep}{0pt} 
    \begin{tabularx}{\linewidth}{
        >{\centering\arraybackslash}X 
        >{\centering\arraybackslash}X    
        >{\centering\arraybackslash}X  
        >{\centering\arraybackslash}X   
        >{\centering\arraybackslash}X}
    \toprule
    \multirow{2}{*}{\textbf{VLM}} 
    & \multicolumn{2}{c}{\textbf{In-Domain}} 
    & \multicolumn{2}{c}{\textbf{Out-Domain}} \\
    \cmidrule(lr){2-3} \cmidrule(lr){4-5}
    & $\mathcal{B}=256$ & $\mathcal{B}=384$ 
    & $\mathcal{B}=256$ & $\mathcal{B}=384$ \\
    \midrule
    LISA (80k) & \textbf{57.4} & \textbf{76.9} & \textbf{60.8} & 77.8 \\
    LISA (55k) & 57.0 & 76.8 & 59.0 & \textbf{78.1} \\
    LISA (24k) & 55.2 & 75.9 & 56.6 & 76.8 \\
    LISA (no fine-tune) & 54.0 & 72.6 & 57.9 & 75.1 \\
    \bottomrule
    \end{tabularx}
    \label{tab:expt-vary-vlm-train-ds}
    \vspace{-5mm}
\end{table}

\begin{table}[t]
\vspace{-0.6cm}
\tiny
\centering
\caption{Scaling up dataset size for fine-tuning CLIP~\cite{stevens2024bioclip} - targets found (\%) $\uparrow$}
\resizebox{\columnwidth}{!}{
\setlength{\tabcolsep}{3pt}
\begin{tabular}{cc ccc ccc @{}p{0.6em}@{} ccc ccc}
\toprule
\multicolumn{2}{c}{} 
    & \multicolumn{6}{>{\columncolor{headergreen}}c}{\textbf{In-domain}}
    & 
    & \multicolumn{6}{>{\columncolor{headergreen}}c}{\textbf{Out-domain}} \\

\addlinespace[1mm]
\multicolumn{2}{c}{\raisebox{0.5\normalbaselineskip}[0pt][0pt]{\textbf{Dataset}}}
    & \multicolumn{3}{c}{\textbf{$\mathcal{B}=256$}}
    & \multicolumn{3}{c}{\textbf{$\mathcal{B}=384$}}
    &
    & \multicolumn{3}{c}{\textbf{$\mathcal{B}=256$}}
    & \multicolumn{3}{c}{\textbf{$\mathcal{B}=384$}} \\
\cmidrule(lr){3-5} \cmidrule(lr){6-8} \cmidrule(lr){10-12} \cmidrule(lr){13-15}
\noalign{\vskip -0.3ex}
\textbf{Size} & \textbf{TTA}
    & {\tiny\textbf{\textit{All}}} & {\tiny\textbf{\textit{Bot. 5\%}}} & {\tiny\textbf{\textit{Bot. 2\%}}}
    & {\tiny\textbf{\textit{All}}} & {\tiny\textbf{\textit{Bot. 5\%}}} & {\tiny\textbf{\textit{Bot. 2\%}}}
    &
    & {\tiny\textbf{\textit{All}}} & {\tiny\textbf{\textit{Bot. 5\%}}} & {\tiny\textbf{\textit{Bot. 2\%}}}
    & {\tiny\textbf{\textit{All}}} & {\tiny\textbf{\textit{Bot. 5\%}}} & {\tiny\textbf{\textit{Bot. 2\%}}} \\
\noalign{\vskip -0.4ex} 
\midrule
380k & Y & 57.4 & \textbf{28.0} & 27.3 & 76.1 & \textbf{53.0} & 51.9 & & 60.8 & 31.7 & \textbf{30.7} & 79.6 & 58.9 & \textbf{56.1} \\

380k & N & 56.6 & \textbf{26.6} & 27.3 & 75.5 & \textbf{50.0} & 51.4 & & 58.5 & 23.1 & \textbf{16.0} & 77.1 & 44.8 & \textbf{36.1} \\

\midrule

200k & Y & 56.5 & 26.0 & \textbf{22.4} & 75.0 & 49.6 & \textbf{46.4} & & 59.8 & 27.7 & \textbf{25.1} & 79.0 & 57.0 & \textbf{55.2} \\

200k & N & 55.6 & 21.7 & \textbf{15.2} & 74.3 & 42.5 & \textbf{32.6} & & 56.1 & 20.3 & \textbf{13.2} & 76.1 & 42.6 & \textbf{37.2} \\

\midrule

80k & Y & 53.7 & 33.0 & \textbf{30.6} & 73.7 & 59.8 & \textbf{58.1} & & 58.7 & \textbf{36.2} & 34.4 & 78.1 & 64.0 & \textbf{62.1} \\

80k & N & 52.8 & 21.7 & \textbf{15.6} & 72.1 & 41.9 & \textbf{38.3} & & 55.7 & \textbf{19.9} & 18.2 & 74.9 & 38.7 & \textbf{31.5} \\

\midrule

No Fine-tune & Y & 49.4 & \textbf{20.9} & 16.0 & 68.4 & \textbf{47.7} & 39.6 & & 50.6 & 18.7 & \textbf{18.6} & 74.0 & 53.9 & \textbf{50.7} \\

No Fine-tune & N & 48.1 & \textbf{19.6} & 17.0 & 67.8 & \textbf{36.5} & 33.5 & & 49.1 & 16.3 & \textbf{11.8} & 69.2 & 34.2 & \textbf{27.7} \\

\bottomrule
\end{tabular}
}
\label{tab:expt-vary-clip-train-ds}
\vspace{-4mm}
\end{table}

\begin{table}[t]
    \scriptsize
    \centering
    \caption{Comparing against prompt learning - targets found (\%) $\uparrow$}
    \begin{tabularx}{\textwidth}{
        >{\centering\arraybackslash}p{0.15\textwidth} 
        >{\centering\arraybackslash}p{0.09\textwidth} 
        >{\centering\arraybackslash}p{0.08\textwidth} 
        >{\centering\arraybackslash}X 
        >{\centering\arraybackslash}X 
        >{\centering\arraybackslash}X 
        >{\centering\arraybackslash}X }
    \toprule
    \multirow{2}{*}{\textbf{Method}} 
    & \multirow{2}{*}{\textbf{LR}} 
    & \multirow{2}{*}{\shortstack{\textbf{Inference}\\\textbf{Time (s)}}} 
    & \multicolumn{2}{c}{\textbf{In-Domain}} 
    & \multicolumn{2}{c}{\textbf{Out-Domain}} \\
    \cmidrule(lr){4-5} \cmidrule(lr){6-7}
    & & & \textbf{$\mathcal{B}$ = 256} & \textbf{$\mathcal{B}$ = 384}
        & \textbf{$\mathcal{B}$ = 256} & \textbf{$\mathcal{B}$ = 384} \\
    \midrule
    \multirow{1}{*}{Weights (Ours)} 
    & (1e-6, 1e-5) & 0.15 
    & \textbf{57.4} & \textbf{76.1} & \textbf{60.8} & \textbf{79.6}  \\  
        
    \multirow{1}{*}{Prompt~\cite{shu2022test}} 
    & (1e-3, 1e-2) & 0.05
    & 57.1 & 75.3 & 59.7 & 78.5  \\  
    
    \midrule
    \multirow{1}{*}{No TTA} 
    & -- & --
    & 56.6 & 75.5 & 58.5 & 77.1  \\      
    \bottomrule
    \end{tabularx}
    \label{tab:expt-prompt-learning}
    \vspace{-0.4cm}
\end{table}

\textbf{Scaling Up CLIP Dataset: }
We measure the performance of our satellite image CLIP encoder fine-tuned with data sets of different sizes in Table~\ref{tab:expt-vary-clip-train-ds}, to justify why we choose to use the full data set of \textbf{380k} images.
In particular, we fine-tune CLIP with images from the full \textbf{380k} dataset, from the \textbf{200k} dataset downsampled from the full dataset, and from the \textbf{80k} AVS dataset (used to generate score maps for VLM fine-tuning).
In addition, we conduct a study in which we do not fine-tune the CLIP model at all. 
We note an increasing trend in search performance as we scale the dataset. 
We use a larger training dataset for our CLIP baselines compared to our VLM baselines as these VLMs already have the added advantages of using CLIP as a foundation and being pretrained on much larger datasets.
We observe a general increase in TTA performance gain when trained on less data. 
In particular, we achieve a TTA performance gain of up to 30.0\% when trained on the 80k dataset ($\mathcal{B}=384$, bot. 2\%).
This indicates Search-TTA's ability to significantly improve score maps when using models trained on less data.

\subsection{AVS Framework Baselines}
\label{app:baselines:avs-frameworks}

We evaluate the effectiveness of the Search-TTA framework by comparing its performance with existing AVS baselines in the remote sensing domain. 
Similar to our setup, VAS~\cite{sarkar2024visual} and PSVAS~\cite{sarkar2023partially} model an AVS problem where an agent, guided by aerial imagery and operating under a fixed query budget, aims to maximize the number of targets found.
VAS utilizes end-to-end reinforcement learning to co-train a feature extraction network and a policy network. 
The detection results gathered during the search process are then piped back into the feature extraction network for prediction updates. 
On the other hand, PSVAS decouples its prediction module from its policy network.
PSVAS pretrains its prediction module using supervised learning and jointly optimizes both modules using reinforcement learning.
During test time, it uses detection results from the search process to directly update the weights of their prediction module. 
Note that their vision backbones (ResNet~\cite{he2016deep}) are not foundation models and must be trained on specific classes.
In addition, both methods do not perform realistic path planning, but instead allow for querying of non-adjacent cells (i.e. teleporting).
For fair comparison, we retain their ability to choose where it wants to query, but also perform Dijkstra path planning to consider the cells on route to their query locations.
We weigh our Dijkstra cost function with a combination of factors, which aims to minimize distance, maximize traveling along paths of high probability (output from their vision module), and avoiding visited cells.

\section{Additional Experiments and Analysis}
\label{app:add-expts}

In this section, we provide information on additional experiments and ablation studies conducted, to supplement the information in Sec.~\ref{sec:experiments}.
Unless mentioned otherwise, we discretize the search space to 24$\times$24 grids, randomize start positions that is consistent across different validation runs, use ground-level images as our query modality, and perform TTA updates every 20 steps or whenever targets are found.

\subsection{Varying TTA Methodologies}
\label{app:add-expts:tta-methods}

\textbf{Prompt Learning:}
We compare our approach to prompt learning~\cite{shu2022test}, where we perform gradient updates on our satellite image prompt instead of our model weights. 
From Table~\ref{tab:expt-prompt-learning}, we noticed that a different learning rate range works better for prompt learning, likely due to the number of parameters that are updated during backpropagation as observed in ~\cite{kaplan2020scaling}.
For fair comparison, we use the same hardware (1x NVIDIA A5000 GPU) to log the inference time.
From our results, we observe that weights fine-tuning achieves better averaged performance but has a slower inference time.
While the number of parameters updated for prompt learning is significantly less, we only observe three times faster in inference speed, likely due to overheads with PyTorch's computational graphs.   
We leave comparison with other fine-tuning methods such as LoRA~\cite{hu2022lora} to future works.

\textbf{Text-based TTA:}
We study the effect of an alternative text-based TTA strategy~\cite{huang2022inner} aside from our SPPP-based strategy.
Instead of integrating the VLMs into our Search-TTA framework which may be time-consuming to test, we design a simple experiment using the region-based statistics logged during our Search-TTA's search process (for \textit{Aves Charadriiformes} / \textit{Shorebirds}).
For each region defined by kmeans clustering, we logged the number of targets found and the ratio of the number of patches explored, at the 25\%, 50\%, 75\%, and 100\% search checkpoints.
We pass these statistics, along with their initial landmark and score predictions (at 0\% checkpoint), into the VLM and prompt them to reconsider their predictions. 
From Table~\ref{tab:expt-text-tta}, we note the inconsistency in RMSE values throughout the different checkpoints, in contrast to the consistent improvements made with our SPPP-based strategy.
This highlights the importance of a principled approach to TTA to achieve consistent results.
We share the prompt design in Appendix~\ref{app:prompt:vlm-baselines}.

\begin{figure}[t]
    \centering
    \includegraphics[width=1.0\linewidth]{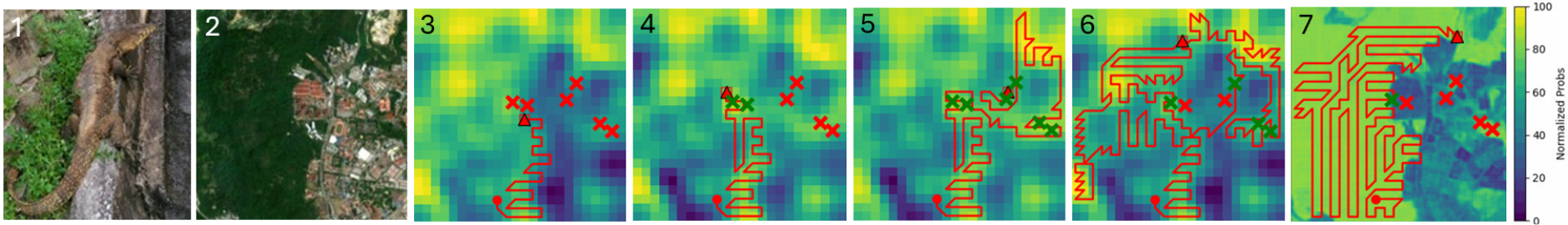}
    \caption{(1) Ground image of a \textit{Monitor Lizard}. (2) Satellite image where \textit{Monitor Lizards} can be found. (3) Initial CLIP probability output. (4) TTA increases probability values in urban region where the two lizards were found. (5) Improved priors leads to efficient search. (6) Inefficient search without TTA. (7) LISA~\cite{lai2024lisa} output probability distribution with higher scores for the forest region, but is unable to perform online adaptation when no targets were found for a prolonged period. }
    \label{fig:tta-vs-lisa-viz}
    \vspace{1mm}
\end{figure}

\begin{table}[t]   \scriptsize
    \centering
    \vspace{-0.4cm}
    \caption{Comparing against text-based TTA (384 Steps) - RMSE (\%) $\downarrow$}
    \begin{tabularx}{\textwidth}{
        >{\centering\arraybackslash}p{0.2\textwidth} 
        >{\centering\arraybackslash}p{0.1\textwidth}
        >{\centering\arraybackslash}X >{\centering\arraybackslash}X >{\centering\arraybackslash}X >{\centering\arraybackslash}X >{\centering\arraybackslash}X}
    \toprule
    \multirow{2}{*}{\textbf{Method}} & \multirow{2}{*}{\textbf{TTA Type}} 
    & \multicolumn{5}{c}{\textbf{\textit{Aves Charadriiformes (Shorebirds)}}} \\
    \cmidrule(lr){3-7}
    & & \textbf{First (0\%)} & \textbf{Quartile (25\%)} & \textbf{Mid (50\%)} & \textbf{Quartile (75\%)} & \textbf{Last (100\%)} \\
    \midrule
    \multirow{2}{*}{CLIP} & SPPP & \textbf{58.7} & \textbf{57.9} & \textbf{56.5} & \textbf{52.4} & \textbf{49.0} \\
                          & -    & 58.7 & 58.7 & 58.7 & 58.7 & 58.7 \\
    \midrule
    \multirow{2}{*}{\shortstack{Qwen2+GroundedSAM\\~\cite{wang2024qwen2,ren2024grounded}}} 
                          & Text & 62.4 & 62.6 & 60.2 & 61.6 & 60.9 \\
                          & -    & 62.4 & 62.4 & 62.4 & 62.4 & 62.4 \\
    \midrule
    \multirow{2}{*}{\shortstack{LLaVA+GroundedSAM\\~\cite{liu2023visual,ren2024grounded}}} 
                          & Text & 59.8 & 59.8 & 60.2 & 58.6 & 62.0 \\
                          & -    & 59.8 & 59.8 & 59.8 & 59.8 & 59.8 \\
    \bottomrule
    \end{tabularx}
    \label{tab:expt-text-tta}
    \vspace{-2mm}
\end{table}

\subsection{Additional Baseline Comparisons}
\label{app:add-expts:baselines}

\textbf{Search-TTA (\textit{In-Domain} Taxonomies):}
In addition to the experimental results presented in Sec.~\ref{sec:expt-tta-planners} (where $\mathcal{B=}$ 256), we present results for $\mathcal{B=}$ 384 in Table~\ref{tab:expt-tta-planners-384steps}.
Similarly, our results show a general improvement in percentage targets found (especially in the bottom percentile), speed of locating the first target, and quality of score maps generated (as measurements are collected during the search process).
This highlights Search-TTA's consistency in improving search performance.

\begin{table}[t] 
\vspace{-5mm} \scriptsize
\centering  
\caption{Evaluating TTA on different planners (CLIP vision model), on \textit{In-domain} taxonomies}
\resizebox{\columnwidth}{!}{
\setlength{\tabcolsep}{3pt}
\begin{tabular}{lccccccc@{}p{0.4em}@{}ccccccc}

\toprule
\multirow{3}{*}{\raisebox{-0.3\normalbaselineskip}[0pt][0pt]{\textbf{Planner Type}}} 
    & \multicolumn{7}{>{\columncolor{headergreen}}c}{\textbf{$\mathcal{B}=256$}} 
    & \multicolumn{1}{c}{} 
    & \multicolumn{7}{>{\columncolor{headergreen}}c}{\textbf{$\mathcal{B}=384$}} \\
\addlinespace[1mm]
    & \multicolumn{3}{c}{\textbf{Found (\%)} $\uparrow$} 
    & \multicolumn{3}{c}{\textbf{RMSE (\%)} $\downarrow$}
    & \multirow{2}{*}{\raisebox{-2.0ex}{\makecell[c]{\textbf{Steps  $\downarrow$}\\\tiny\textbf{(First tgt)}}}}
    & 
    & \multicolumn{3}{c}{\textbf{Found (\%)} $\uparrow$} 
    & \multicolumn{3}{c}{\textbf{RMSE (\%)} $\downarrow$}
    & \multirow{2}{*}{\raisebox{-2.0ex}{\makecell[c]{\textbf{Steps  $\downarrow$}\\\tiny\textbf{(First tgt)}}}} \\
\cmidrule(lr){2-4} \cmidrule(lr){5-7} \cmidrule(lr){10-12} \cmidrule(lr){13-15}
\noalign{\vskip -0.4ex} 
    & {\tiny\textbf{\textit{All}}} & {\tiny\textbf{\textit{Bot. 5\%}}} & {\tiny\textbf{\textit{Bot. 2\%}}}
    & {\tiny\textbf{\textit{First}}} & {\tiny\textbf{\textit{Mid}}} & {\tiny\textbf{\textit{Last}}}
    & 
    & 
    & {\tiny\textbf{\textit{All}}} & {\tiny\textbf{\textit{Bot. 5\%}}} & {\tiny\textbf{\textit{Bot. 2\%}}}
    & {\tiny\textbf{\textit{First}}} & {\tiny\textbf{\textit{Mid}}} & {\tiny\textbf{\textit{Last}}}
    & \\
\noalign{\vskip -0.3ex} 

\midrule
RL (TTA)~\cite{cao2023ariadne} 
    & 57.4 & \textbf{28.0} & 27.3 
    & 54.4 & 53.7 & \textbf{51.0} 
    & 85.1
    & 
    & 76.1 & \textbf{53.0} & 51.9 
    & 54.4 & 53.0 & \textbf{45.9} 
    & 101.9 \\

RL (no TTA)~\cite{cao2023ariadne} 
    & 56.6 & \textbf{26.6} & 27.3 
    & 54.4 & 54.4 & \textbf{54.4} 
    & 84.8
    & 
    & 75.5 & \textbf{49.9} & 51.4 
    & 54.4 & 54.4 & \textbf{54.4} 
    & 102.7 \\

\midrule

IS (TTA)~\cite{lanillos2014multi} 
    & 52.1 & 28.1 & \textbf{22.1}
    & 54.4 & 53.4 & \textbf{51.3}  
    & 90.5
    & 
    & 71.0 & 45.0 & \textbf{40.8} 
    & 54.4 & 52.9 & \textbf{46.2} 
    & 109.6 \\

IS (no TTA)~\cite{lanillos2014multi} 
    & 51.2 & 21.8 & \textbf{14.9} 
    & 54.4 & 54.4 & \textbf{54.4} 
    & 90.8
    & 
    & 70.2 & 39.7 & \textbf{35.3} 
    & 54.4 & 54.4 & \textbf{54.4} 
    & 111.2 \\

\midrule

Lawnmower~\cite{choset2001coverage} 
    & 41.8 & -- & -- 
    & -- & -- & -- 
    & 112.9
    & 
    & 71.5 & -- & -- 
    & -- & -- & -- 
    & 148.3 \\
    
\bottomrule
\end{tabular}
}
\label{tab:expt-tta-planners-384steps}
\vspace{-6mm}
\end{table}

\textbf{Varying Vision Model:}
We provide more insights to the data presented in Sec.~\ref{sec:expt-baselines} (Table~\ref{tab:expt-vision-models}).
We note the significantly longer inference speed for the Qwen2-VL+GroundedSAM and LLaVA+GroundedSAM baselines.
This is because, unlike the other VLMs, Qwen2-VL and LLaVA are required to output landmark names and scores in text, which involves significantly more token generation compared to the custom \textit{[SEG]} token used in LISA and LLM-Seg.
LLaVA is slower compared to Qwen2-VL because we use a LLaVA-13B model compared to the Qwen2-VL-7B model. 
Note that we take all inference speed measurements on a single NVIDIA A5000 GPU.

In addition, we provide snapshots of Search-TTA and LISA to compare their performance when searching for \textit{Monitor Lizards} (\textit{reptilia Aquamata Varanidae}).
From Figure~\ref{fig:tta-vs-lisa-viz}, we can see that Search-TTA increases probability values in the urban region after collecting positive samples there. 
This online adaptation results in a more efficient search (6 targets found). In contrast, LISA over-exploits its initial belief where \textit{Monitor Lizards} are more likely to be found in the forest region, and is unable to correct its probability distribution despite many negative measurements (only 1 target found).
Note that our approach without TTA results in only 4 targets found.

\textbf{AVS Baseline:}
In addition to the experimental results presented in Sec.~\ref{sec:expt-multimodal-inputs} (\textit{Animalia Chordata Aves Charadriiformes}), 
we present results for the same AVS framework baselines when searching for \textit{Animalia Chordata Reptilia Squamata} in Table~\ref{tab:expt-search-frameworks-v2}.
Likewise, we notice the same trend where Search-TTA outperforms almost all baselines in terms of percentage targets found.
This highlights Search-TTA's versatility, given that it is able to outperform AVS baselines (pretrained on specific taxonomies) with just a single model. 
Although Lawnmower outperforms VAS and PSVAS when $\mathcal{B}=384$, VAS and PSVAS find the first target more quickly by performing a more targeted search.

\begin{table}[t]   \scriptsize
    \centering
    \caption{Comparing AVS frameworks (\textit{Animalia Chordata Reptilia Squamata})} 
    \begin{tabularx}{\textwidth}{>{\centering\arraybackslash}p{0.15\textwidth}
                        >{\centering\arraybackslash}X >{\centering\arraybackslash}X >{\centering\arraybackslash}X 
                        >{\centering\arraybackslash}X >{\centering\arraybackslash}X >{\centering\arraybackslash}X }
    \toprule
    \multirow{2}{*}{\textbf{Frameworks}} 
    & \multicolumn{3}{c}{\textbf{$\mathcal{B}$ = 256}} 
    & \multicolumn{3}{c}{\textbf{$\mathcal{B}$ = 384}} \\
    \cmidrule(lr){2-4} \cmidrule(lr){5-7}
    & \shortstack[c]{\textbf{Found}\\\textbf{(\%)}} 
    & \shortstack[c]{\textbf{Explored}\\\textbf{(\%)}} 
    & \shortstack[c]{\textbf{Steps}\\\textbf{(First tgt)}} 
    & \shortstack[c]{\textbf{Found}\\\textbf{(\%)}} 
    & \shortstack[c]{\textbf{Explored}\\\textbf{(\%)}} 
    & \shortstack[c]{\textbf{Steps}\\\textbf{(First tgt)}} \\
    \midrule
    CLIP+RL (TTA) & \textbf{60.3} & 44.3 & 86.9 & \textbf{80.5} & 65.6 & 91.3 \\
    CLIP+RL (no TTA) & 55.9 & 44.3 & 78.7 & 76.9 & 65.6 & 95.7 \\
    PSVAS~\cite{sarkar2023partially} & 47.3 & 43.1 & 83.7 & 70.0 & 62.1 & 109.0 \\
    VAS~\cite{sarkar2024visual} & 46.8 & 44.1 & \textbf{75.5} & 68.5 & 64.6 & \textbf{89.3} \\
    Lawnmower~\cite{choset2001coverage} & 43.5 & \textbf{44.4} & 116.0 & 71.4 & \textbf{66.7} & 149.1 \\
    \bottomrule
    \end{tabularx}
    \label{tab:expt-search-frameworks-v2}
    \vspace{-0.7cm}
\end{table}

\textbf{Experimental Validation:}
In addition to the details for our AVS hardware-in-the-loop experiments described in Sec.~\ref{sec:expt-real-robot}, we also rescale our Yosemite Valley simulated environment from 865m $\times$ 865m to 280m $\times$ 280m. 
This is due to the crazyflie's limited flight time of 5 minutes, which is not indicative of larger drones that often have longer battery life.
The simulated drone flies at an altitude of 30m and is equipped with a $\text{90}^\circ$ FOV camera mounted on a gimbal (tilted at $\text{30}^\circ$ to achieve better bear detection rates). 
We conduct both experiments, with and without TTA, using $\mathcal{B}=300s$ (traveled 2058m and 2062m respectively).
We execute TTA every 5 iterations when it is enabled.

\subsection{Time Analysis on Different Hardware}
\label{app:add-expts:time-analysis}

\setlength{\intextsep}{4.5pt} 
\begin{wraptable}{r}{0.5\textwidth} 
    \centering
    \scriptsize
    \caption{Time analysis on different hardware} 
    \vspace{-2mm}
    \setlength{\tabcolsep}{0pt} 
    \begin{tabularx}{\linewidth}{
        >{\raggedright\arraybackslash\hspace{0pt}}p{0.25\linewidth}
        >{\centering\arraybackslash}X    
        >{\centering\arraybackslash}X
        >{\centering\arraybackslash}X}
    \toprule
    \textbf{Stage} & \textbf{Frequency} & \textbf{RTX A5000} & \textbf{Orin AGX} \\
    \midrule
    CLIP (Inference) & Once & 0.11s & 0.14s \\
    CLIP (TTA) & Every 20s & 0.15s & 0.37s \\
    Kmeans & Once & 0.11s & 0.97s \\
    \bottomrule
    \end{tabularx}
    \label{tab:search-tta-time-analysis}
\end{wraptable}

We measure the time taken for each stage of the Search-TTA framework on the server (NVIDIA RTX A5000, AMD Ryzen threadripper 3970x) and on an edge device (NVIDIA Orin AGX), and report these measurements in Table~\ref{tab:search-tta-time-analysis}.
Note that CLIP inference and Kmeans clustering only needs to be perfored once before the search episode.
Although TTA performs slower on an edge device, the search planner can fall back on previous TTA-corrected heatmaps since it does not need to rely on the latest heatmap in a time-sensitive manner.
Moreover, in typical drone monitoring applications, a datalink is established between the drone and a more powerful ground control station (on which we can perform TTA).
Note that the time taken for the SPPP loss calculation is negligible, as is the increase in time taken for TTA with more measurements; we also always perform a single backpropagation step per TTA iteration.
However, the time taken for TTA will scale with VLM size given that we fine-tune the full model.
In the future, we will explore Parameter-Efficient Fine-Tuning~\cite{hu2022lora} techniques to selectively perform weight updates and learning-based clustering methods to achieve efficient TTA and clustering of high-dimensional embeddings.

\section{Prompt Engineering}
\label{app:prompt}

\subsection{Score Map Generation Prompts}
\label{app:prompt:gpt4o-score-maps}

\begin{PromptBlock}{GPT4o Relabelling Prompt}
\begin{Verbatim}[
  fontsize=\scriptsize,
  breaklines=true,
  breakanywhere=true,
  breaksymbolleft={}
]
You are an AI visual assistant that can analyze a single satellite image that is very zoomed out (covering around 2 km over the width and 2 km over the height of the image). You are given the bounding box of the segmentaions of different regions, represented as (x1, y1, x2, y2) with floating numbers ranging from 0 to 1. These values correspond to the top-left x, top-left y, bottom-right x, and bottom-right y. These values correspond to the x and y coordinates. x-coordinates increase from left to right, and y-coordinates increase from top to bottom. The closer the instance is to the top edge, the smaller the y-coordinate. Assume that point1 [0, 0] is at the upper left, and point2 [1, 1] is at the bottom right. 

The image contains different regions defined by bounding boxes. In the input data the regions are numbered from 0 to 3. The region names are: "Urban Area", "Barren", "Water", "Vegetation". Your task is to verify if the region inside the bounding box seems correct as seen in the image. Generate a conversation between yourself (the AI assistant) and a user asking about the photo. Verify the region names for each region given in the input data and tell me if it's name matches visually as seen from the image. When a region is incorrect then also remap the region name to the correct landmark name. Your responses should be in the tone of an AI that is "seeing" the image and answering accordingly. 

When using all the provided information, directly generate the conversation. Always answer as if you are directly looking at the image.

Mapping:
{
    "Region 0": "Urban Area",
    "Region 1": "Barren",
    "Region 2": "Water",
    "Region 3": "Vegetation"
}

You must return your response in the JSON format:  
{  
    "conversation": [  
        {  
            "from": "human",  
            "sat_key": sat_key,  
            "taxonomy": taxonomy,  
        },  
        {
            "from": "gpt",
            "landmarks": {
                "Region i": "Correct/Incorrect",
                "Region j": "Correct/Incorrect",
                "Region k": "Correct/Incorrect",
            },
            "corrected_landmarks": {
                "Region i": {
                    "name": "Urban Area/Barren/Water/Vegetation",
                },
                "Region j": {
                    "name": "Urban Area/Barren/Water/Vegetation",
                },
                "Region k": {
                    "name": "Urban Area/Barren/Water/Vegetation",
                },
            }
        }
 
    ]  
}

{Examples}

{Input_Data}

Double check if the region names are correct. In the answer if the region name seems incorrect then remap it to the correct landmark name, else leave it as it is. Once again, please output your response in the JSON format only.

\end{Verbatim}
\end{PromptBlock}

\vspace{1em}

\begin{PromptBlock}{GPT4o Scoring Prompt}
\begin{Verbatim}[
  fontsize=\scriptsize,
  breaklines=true,
  breakanywhere=true,
  breaksymbolleft={}
]

You are an AI visual assistant that can analyze a single satellite image that is very zoomed out (covering around 2 km over the width and 2 km over the height of the image). A specific animal/plant location within the image is given, along with detailed coordinates. The locations are in the form of coordinates, represented as (x,y) with floating numbers ranging from 0 to 1. These values correspond to the x and y coordinates. x-coordinates increase from left to right, and y-coordinates increase from top to bottom. The closer the instance is to the top edge, the smaller the y-coordinate. Assume that point1 [0, 0] is at the upper left, and point2 [1, 1] is at the bottom right.  

You are also given the name of the animal/plant as taxonomy. Along with the exact coordinates in the image, you are also given the bounding box of the area where the animal/plant was found, represented as (x1, y1, x2, y2) with floating numbers ranging from 0 to 1. These values correspond to the top-left x, top-left y, bottom-right x, and bottom-right y. The values follow the same format as the coordinates.  

The image contains different landmarks defined by bounding boxes. If a coordinate lies inside a particular bounding box and that bounding box belongs to a specific landmark, that means the animal/plant was found in that landmark. Your task is to generate a conversation between yourself (the AI assistant) and a user asking about the photo. Your responses should be in the tone of an AI that is "seeing" the image and answering accordingly. Using the image, taxonomy, coordinates, and bounding box information, provide a score for each landmark and a detailed explanation of where the animal/plant could be found among those landmarks based on the landmark semantics from the image. Add a simple question in the conversation, asking about where to find the particular animal/plant in the image (use common name as well as taxonomy for question).

Given the explanation, evaluate a score based on the likelihood of finding the queried animal/plant in each landmark. The length of the 'landmarks' list must match the length of the 'score' list. Probability scores must range between 0.0 and 1.0. It is acceptable for multiple consecutive landmarks to have the same scores. For example, a frog may have the same score for both water and land-type landmarks. Scores can be any value between 0.0 and 1.0, such as 0.1, 0.3, 0.5, 0.7, or 0.9. You must be more conservative, where if you are not sure, you should assign a lower score. If the animals cannot exist inside this landmark, please assign a score of 0.0. For example, a land animal or a non-aquatic plant cannot live inside the water body.

The scoring system is defined as follows:  

1.0: Almost guaranteed to find the animal in the landmark  
0.8: Very likely to find the animal in the landmark  
0.6: Likely to find the animal in the landmark  
0.4: Unlikely to find the animal in the landmark  
0.2: Very unlikely to find the animal in the landmark  
0.0: Almost impossible to find the animal in the landmark  

You must answer with an explanation, landmarks, and scores for each region. Score the area of a particular region not only based on the number of targets actually present but also using the semantic information of the region. Provide a detailed explanation for your scoring, describing the relationship between the region's semantics and the animal/plant.

When using all the provided information, directly generate the conversation. Always answer as if you are directly looking at the image. Do not mention bounding box or region numbers explicitly, instead use the assigned landmark names. Only output the landmark name, not the landmark coordinates. Your answer may include multiple landmark types. It is acceptable for multiple consecutive landmarks to have the same scores.

You must return your response in the JSON format:  

{  
    "conversation": [  
        {  
            "from": "human",  
            "sat_key": sat_key,  
            "taxonomy": taxonomy,  
            "common_name": common_name,
            "question": question using taxonomy,
        },  
        {
            "from": "gpt",
            "explanation": answer,
            "landmarks": {
                "landmark1": {
                    "score": score,
                },
                "landmark2": {
                    "score": score,
                },
                "landmark3": {
                    "score": score,
                },
            }
        }
    ]  
}

{Examples}

{Input_Data}

Once again, please output your response in the JSON format.

\end{Verbatim}
\end{PromptBlock}

\subsection{VLM Baselines Prompts}
\label{app:prompt:vlm-baselines}

\begin{PromptBlock}{Llava Inference Prompt (LLaVA+GroundedSAM Baseline)}
\begin{Verbatim}[
  fontsize=\scriptsize,
  breaklines=true,
  breakanywhere=true,
  breaksymbolleft={}
]
Using the image as a reference, where can {animal} be found? Give me 1-2word high-level landmark names where it can be found in the image. Your response will be a json object where the landmark names are the keys and the probability of it being found in the landmark as values e.g {{"barren_land": 0.6}}. Return just between 3 to 5 landmarks. Your response must be a single json object enclosed in double quotes without additional text.
\end{Verbatim}
\end{PromptBlock}

\vspace{1em}

\begin{PromptBlock}{Llava TTA Prompt (LLaVA+GroundedSAM Baseline)}
\begin{Verbatim}[
  fontsize=\scriptsize,
  breaklines=true,
  breakanywhere=true,
  breaksymbolleft={}
]
You are provided with a heat map of the satellite image earlier, region statistics showing the percentage of the region in the satellite image that has been explored and number of {animal} found in these regions. Region Statistics:\n{explore_info}\n Use these information to update the probabilities of {animal} being found in the landmarks generated previously: {orig_response}. Do not associate the region Rn with any of the landmarks, they are not related in any way. Return your answer as a JSON object in this format: {new_dict}, where the keys are enclosed with double quotes. Begin your answer with explanation and reasoning steps for calculating the new probability values for {landmark_names}, then return the JSON object. You must enclose the JSON object within ```json tag. e.g '''json{{"<landmark>": <new value>}}```.
\end{Verbatim}
\end{PromptBlock}

\vspace{1em}

\begin{PromptBlock}{Qwen Inference Prompt (Qwen2+GroundedSAM Baseline)}
\begin{Verbatim}[
  fontsize=\scriptsize,
  breaklines=true,
  breakanywhere=true,
  breaksymbolleft={}
]
Using the image as a reference, where can {animal} be found? Give me 1-2word high-level landmark names where it can be found in the image. Your response will be a json object where the landmark names are the keys and the probability of it being found in the landmark as values e.g {{"barren_land": 0.8}}. Return just between 3 to 5 landmarks. The key should be landmark names, not any other animals or food. Your response must be a single json object enclosed in double quotes without additional text, and do not return the examples given as a result.
\end{Verbatim}
\end{PromptBlock}

\vspace{1em}

\begin{PromptBlock}{Qwen TTA Prompt (Qwen2+GroundedSAM Baseline)}
\begin{Verbatim}[
  fontsize=\scriptsize,
  breaklines=true,
  breakanywhere=true,
  breaksymbolleft={}
]
You are provided with a satellite image,a heat map of the satellite image and region statistics and tasked to use these information to come up with pairs of landmark:probabilities, where probabilites denotes the chances of finding {animal} in the landmark. The heat map, together with region statistics shows the percentage of the region in the satellite image that has been explored, as well as the number of {animal} found in it. Region Statistics:\n{explore_info}\nSuppose the original response {orig_response}, you need to use the heat map and region statistics information to come up with the new probabilities associated with the given landmark, using the satellite image as a reference. Do not associate regions Rn with {landmark_names}, they are not related to one another. Return your answer in this format: {new_dict} with new probability values (between 0 and 1) you calculated from the region statistics. Begin your response with explanation and reasoning steps for calculating new values for {landmark_names}, and return a single JSON object in {new_dict} format with the new probability values, enclosed in double quotes for the key and values. Enclose the json object within the ```json tag. e.g ```json{new_dict}```. The JSON key should be enclosed by double quotes.
\end{Verbatim}
\end{PromptBlock}

\end{document}